\documentclass{article}

\usepackage{graphicx,amsmath,amssymb}
\newcommand{\be}{\begin{equation}}
\newcommand{\ee}{\end{equation}}
\usepackage[margin=1.0in]{geometry}
\usepackage{xcolor}

\usepackage{eucal}

\usepackage[english]{babel}
\usepackage[utf8x]{inputenc}
\usepackage[T1]{fontenc}
\usepackage{enumitem}

\usepackage{amsmath}
\usepackage{graphicx}
\usepackage{subcaption}

\usepackage{float}
\usepackage{graphicx}
\usepackage{epstopdf, epsfig}
\usepackage{flushend}
\usepackage{latexsym}
\usepackage{caption}
\usepackage{color,soul}
\usepackage{physics}
\usepackage[normalem]{ulem}

 \usepackage{physics}
 \usepackage{amssymb}
 \usepackage{amsmath}

\usepackage{afterpage}

\graphicspath{{figures/}}
\usepackage[colorlinks=true, allcolors=blue]{hyperref}
\usepackage{xcolor}
\usepackage{pgfplots}
\usepackage{bm}
\usepackage{todonotes}
\usepackage{comment}
\usepackage{authblk}

\usepackage{lineno}

\newcommand{\ethan}[1]{\textcolor{black}{#1}}

\usepackage{algorithm,amsmath,algpseudocode,algcompatible,setspace}

\definecolor{light-yellow}{rgb}{1,1,0.75}

\title{Discovering and forecasting extreme events via active learning in neural operators}
    \author[1,\footnote{Co-corresponding authors: pickering@mit.edu, sapsis@mit.edu}]{Ethan Pickering}
    \author[1]{Stephen Guth}
    \author[2]{George Em Karniadakis}
    \author[1,$^*$]{Themistoklis P. Sapsis}

    \affil[1]{Department of Mechanical Engineering, Massachusetts Institute of Technology, Cambridge, MA 02138, USA}
    \affil[2]{Division of Applied Mathematics, Brown University, Providence, RI 02906, USA}

\begin{document}
    \maketitle
    \begin{abstract}
    Extreme events in society and nature, such as pandemic spikes, rogue waves, \ethan{or structural failures}, can have catastrophic consequences. Characterizing extremes is difficult as they occur rarely, arise from seemingly benign conditions, and belong to complex and often unknown infinite-dimensional systems.  Such challenges render attempts at characterizing them as moot. We address each of these difficulties by combining novel training schemes in Bayesian experimental design (BED) with an ensemble of deep neural operators (DNOs). This model-agnostic framework pairs a BED scheme that actively selects data for quantifying extreme events with an ensemble of DNOs that approximate infinite-dimensional nonlinear operators. We find that not only does this framework clearly beat Gaussian processes (GPs) but that 1) shallow ensembles of just \textit{two} members perform \textit{best}; 2) extremes are uncovered regardless of the state of initial data (i.e. with or without extremes); 3) our method eliminates ``double-descent'' phenomena; 4) the use of batches of suboptimal acquisition samples compared to step-by-step global optima does not hinder BED performance; and 5) Monte Carlo acquisition outperforms standard optimizers in high-dimensions. Together these conclusions form the foundation of an AI-assisted experimental infrastructure that can efficiently infer and pinpoint critical situations across many domains, from physical to societal systems.
    
    \end{abstract}

\section{Introduction}


The grand challenge of predicting disasters remains an extremely difficult and unsolved problem \cite{NRC2005}. Disasters, such as pandemic spikes, \ethan{structural failures,} wildfires, or rogue waves\footnote{Rogue waves are rare, giant waves that pose a danger to ships and offshore structures. The largest wave on record was 25.6 meters and hit the Draupner oil platform in the North Sea on January 1, 1995 \cite{hansteen2003observed}. The ``most extreme'' wave, three times the size of surrounding waves, was 19.5 meters and observed off the coast of Ucluelet, British Columbia, on November 17, 2020 \cite{gemmrich2022generation}}, are uniquely challenging to quantify. This is because they are both \textit{rare} and arise from an \textit{infinite} set of physical conditions \cite{sapsis2021statistics}. The proposition of predicting extremes is analogous to \textit{finding a catastrophic needle in an infinite-dimensional haystack}. This calls for methods that can both \textit{discover} extreme events and encode \textit{physical} phenomena into their modeling strategy. We present a Bayesian-inspired experimental design (BED) approach, described in detail in figure \ref{fig:ExpDesignSchematicDisc}, that addresses both challenges by combining a probabilistic ``discovery'' algorithm \cite{blanchard2021output, blanchard2021bayesian} with a deep neural operator designed to approximate physical systems \cite{lu2021learning}.  

Discovery of extremes is often simplified by distilling complex systems to their governing input variables and relevant output variables. Within this interpretation, quantification of extremes has historically taken the form of importance sampling, which uses a biasing distribution to identify regions of the input space that exhibit extreme values \cite{kahn1953methods,shinozuka1983basic}. Unfortunately, these techniques often require additional and challenging considerations for accurate results \cite{dematteis2019extreme,uribe2021cross,wahal2019bimc} and are static, lacking an ability to adjust to new information gained through experiments. Active learning, and specifically BED, provides a dynamic approach that learns from acquired data before selecting new and intriguing input-output data. 

\ethan{Active learning (AL) refers to a broad class of sequential sampling techniques for assembling efficient training datasets. AL has been applied with neural networks in several fields, predominantly in classification tasks such as image recognition \cite{gal2017deep}, text recognition \cite{zhang2017active}, or object detection \cite{aghdam2019active} (see \cite{ren2021survey} for a survey of similar AL applications), with less attention in the literature on regression of physical processes, let alone rare events. While there are some exceptions for AL in rare-event quantification, such as combining DNNs \cite{xiang2020active} or other surrogate models \cite{ehre2022sequential} with weighted importance sampling for structural reliability analyses, neither leverage uncertainty to ensure the input space has been adequately explored. On the other hand, techniques employing BED and uncertainty predictions via Kriging or Gaussian Process (GP) regression \cite{echard2011ak, blanchard2021output} perform well, but cannot be applied to infinite-dimensional systems or scale to large-training sets. This necessitates a solution that can 1) accurately generalize to infinite-dimensional systems and easily scale with data size, 2) emit uncertainty estimates, and 3) apply appropriately defined acquisition functions for selecting extreme data.}


 



Deep neural operators (DNO), such as DeepONet \cite{lu2021learning}, are built specifically for handling infinite-dimensional systems and provide the ideal surrogate model for characterizing extremes. Unlike other ML approaches, such as GPs, which map parameterizations of physical phenomenon, DNOs directly map \textit{physical}, infinite-dimensional functions to \textit{physical}, infinite-dimensional functions. This leads to drastic improvements in generalization to unseen data in high dimensions. Additionally, the neural network backbone of DNOs mean they are intrinsically amenable to big data, unlike GPs which scale as the third power of data size \cite{snelson2006sparse,titsias2009variational}. However, DNOs utility for Bayesian experimental design is an open question as DNOs do not explicitly provide a measure of uncertainty. We propose and show the efficacy of using an ensemble of DNOs for uncertainty quantification and BED. Although much of the literature is skeptical of the generality of ensembles to provide uncertainty estimates, recent viewpoints, \cite{pickering2022structure} and notably \cite{wilson2020bayesian}, have argued that \ethan{deep neural network} (DNN) ensembles provide a very good approximation of the posterior. Our results support this perspective. 

Appropriately defined acquisition functions for uncovering extreme behavior are just as critical as the chosen surrogate model. Recently, \cite{sapsis2020output, blanchard2021output}, in concert with several other works \cite{mohamad2018sequential} and \cite{blanchard2021bayesian}, introduced a class of probabilistic acquisition functions specifically  designed for quantifying extreme events under asymptotically optimal conditions \cite{sapsis22}. By combining statistics of the input space along with statistics deduced from the surrogate model, the method can account for the importance of the output relative to the input. This approach significantly reduces the number of input samples required to characterize extreme phenomena.

\ethan{The main contribution of this work is a scalable, Bayesian-inspired DNO framework with extreme acquisition functions that efficiently learns to discover and forecast extreme events. This framework comes with several favorable properties, such as improved performance to other approaches, computational tractability, robustness, and ease of implementation. The most significant are summarized below:
\begin{enumerate}
    \item The DNO framework consistently beats Gaussian processes (GPs) approaches, especially as dimensionality increases.
    \item Shallow ensembles of just \textit{two} members perform \textit{best}, greatly reducing the training cost of ensemble DNOs.
    \item The use of batches of suboptimal acquisition samples compared to step-by-step global optima does not hinder DNO-BED performance, permitting parallel experimentation in real-life application.
    \item Extremes are uncovered regardless of the state of initial data (i.e. with or without extremes).
    \item The method is observed to eliminate ``double-descent'' phenomena.
\end{enumerate}
Equipped with both extreme acquisition functions and an ensemble of DNOs, our study demonstrates the above contributions through testing three classes of representative high-dimensional systems to discover extreme rogue waves, pinpoint dangerous pandemic scenarios, and efficiently estimate structural stresses to inform ship design.}


\section{Results}
 

\begin{figure} 
\scriptsize{} $a)$ \scriptsize{Pinpoint Dangerous Pandemic Spikes} \hspace{0.15cm} $b)$  \scriptsize{Discover and Predict Rogue Ocean Waves}  \hspace{0.15cm} $c)$ \scriptsize{Estimate Peak Structural Ship Stress} \vspace{0.05cm} \\ 
\includegraphics[width=0.33\textwidth,trim={7cm 0cm 8cm 0cm},clip]{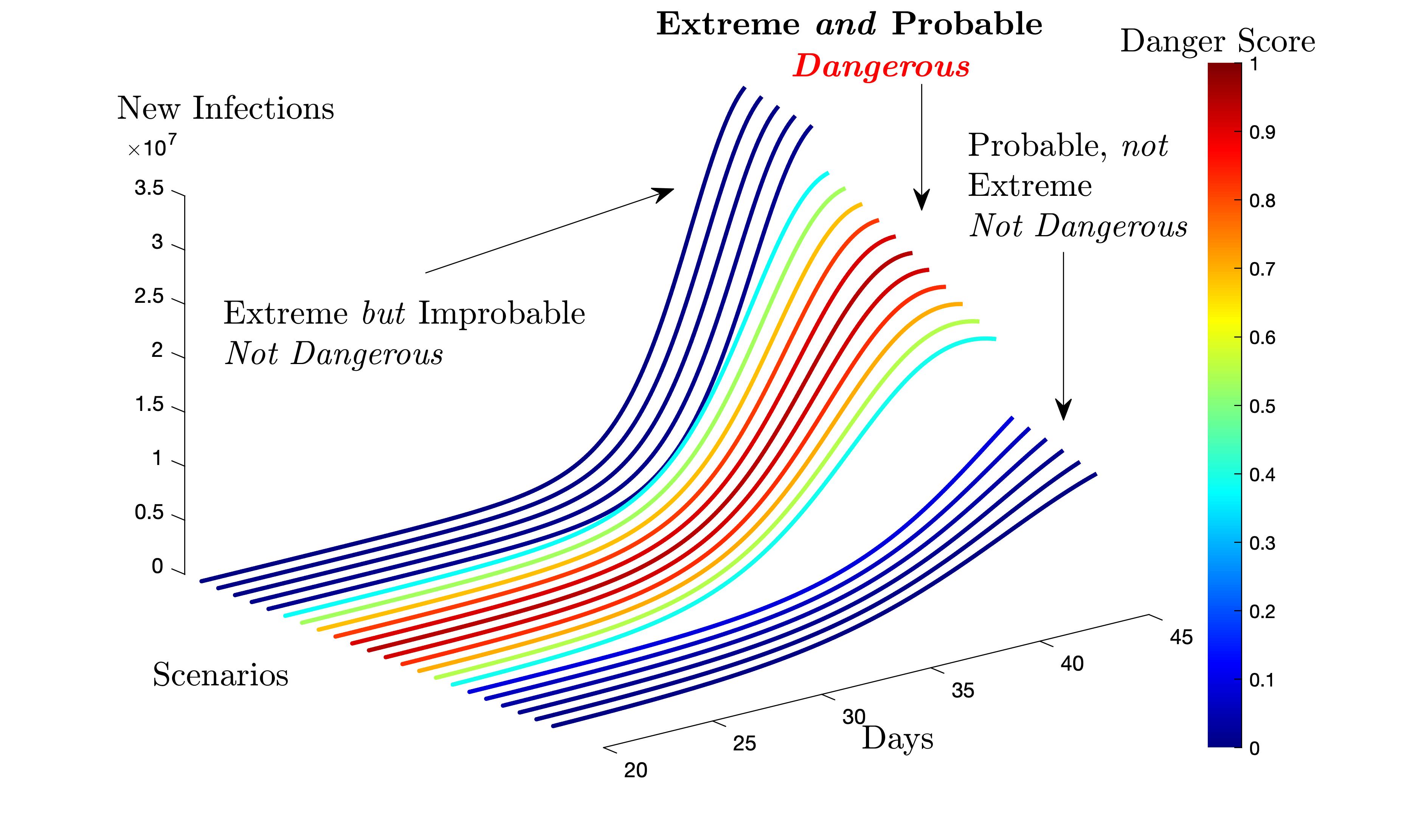}
\includegraphics[width=0.33\textwidth,trim={7cm 0cm 8cm 0cm},clip]{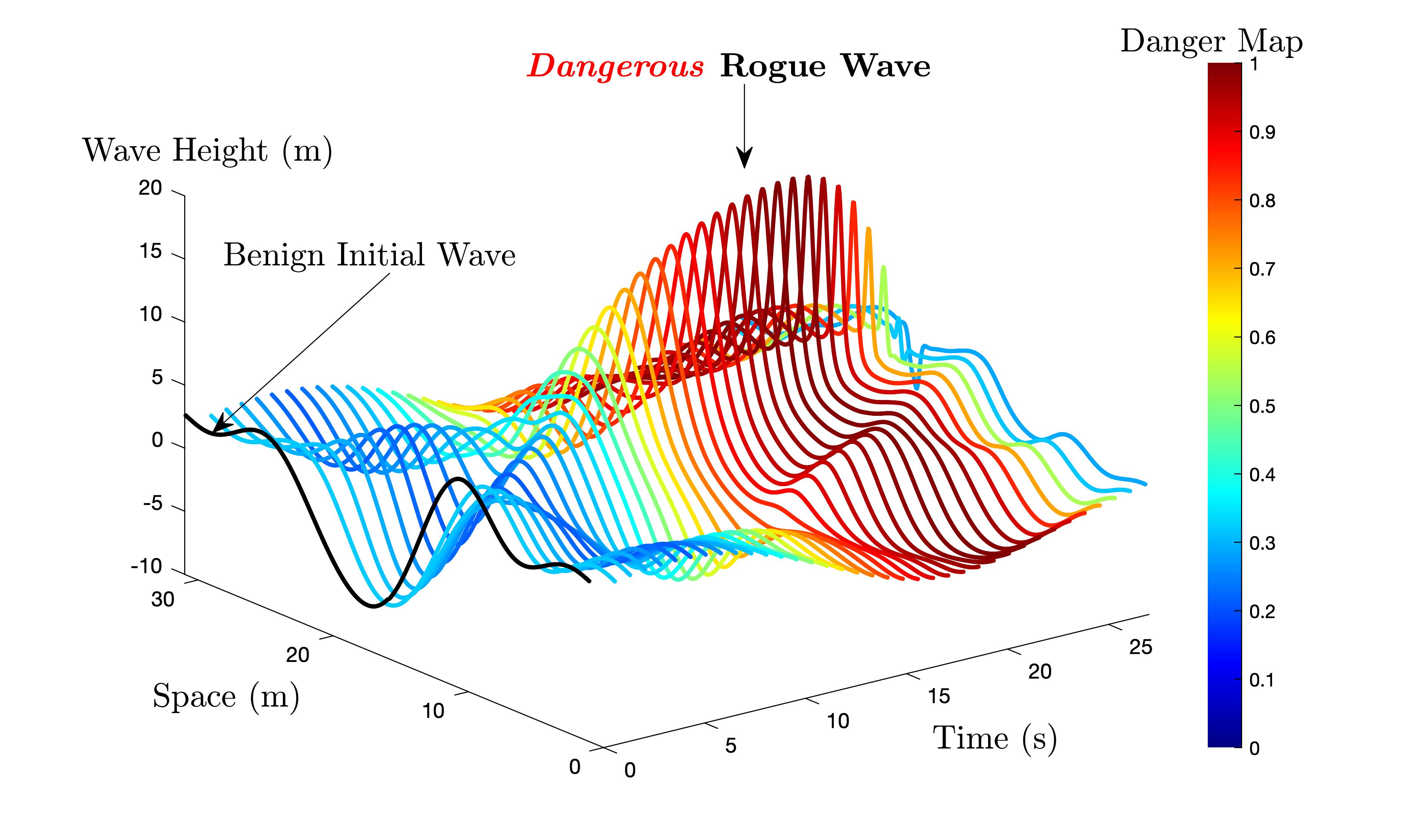} \includegraphics[width=0.33\textwidth,trim={0cm 0cm 0cm 0cm},clip]{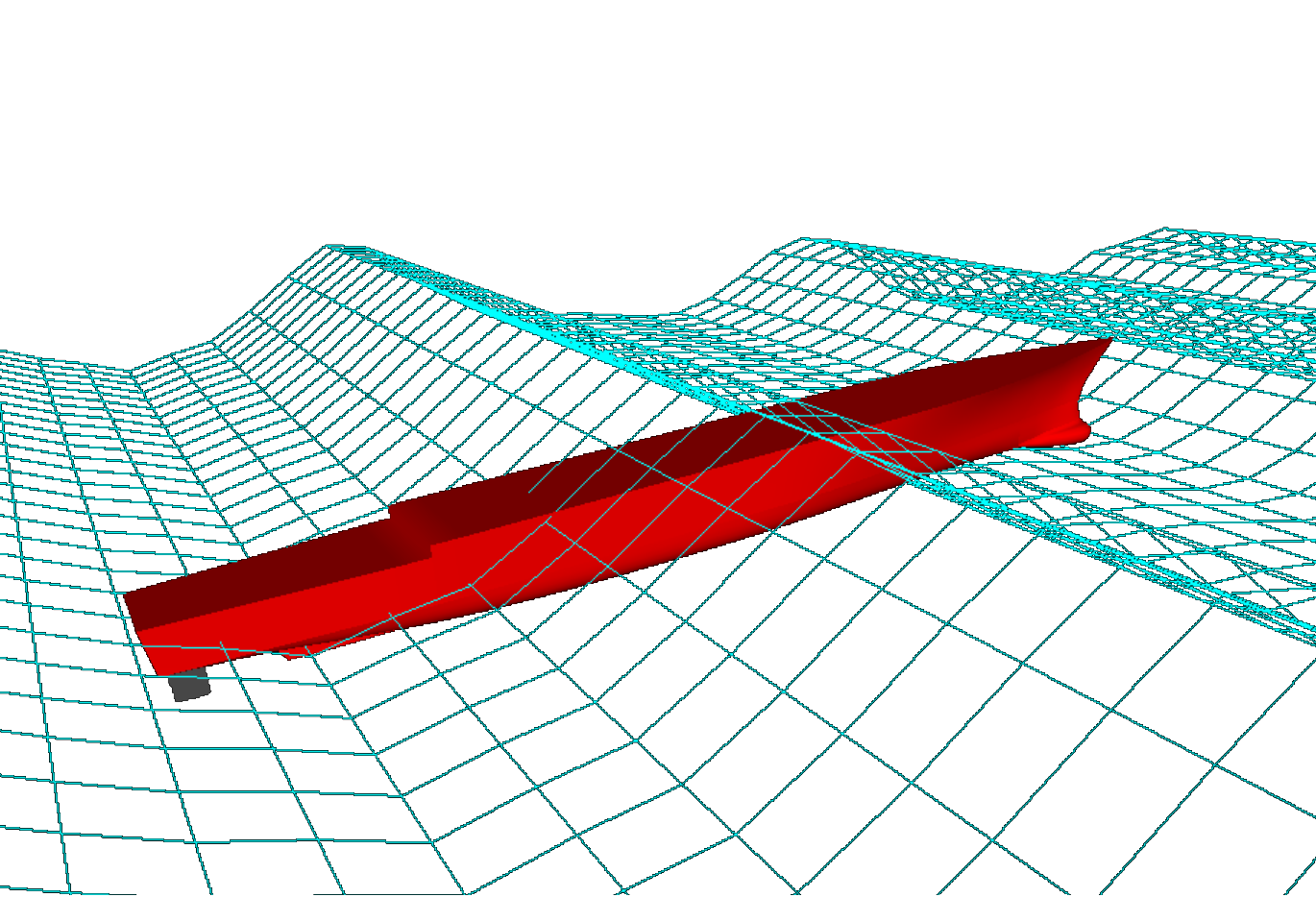}
\caption{\textbf{Inference of diverse extreme phenomena, from pandemic spikes to rogue ocean waves to large structural stress events on ships.} Our framework pinpoints the most dangerous (i.e., probable and extreme) pandemic scenarios (left), i.e. realizations of stochastic infection rates, discovers rogue waves (center), and efficiently estimates large structural stress events for reliable ship design. $a)$ The most dangerous pandemic scenarios are pinpointed by inferring the number of new infections in time for a plurality of infection rate hypotheses. See figure \ref{fig:PandemicExample} for more details. $b)$ Rogue waves are discovered and quantified for future prediction by uncovering the probable wave conditions that non-linearly interact in time to emit rogue waves over three times their original size. We show one example of this phenomena here and refer to figure \ref{fig:Lines_LHS} for more details on discovering these waves. $c)$ The statistics of peak stress govern fatigue lifetimes, with our approach we can efficiently estimate how unique ship designs structurally react to stochastic ocean waves to inform reliable and safe ship design. See figure \ref{fig:LAMP_Learning_Curves} for the stress state related to this graphic.}
\label{fig:Rogue_Pandemic}
\end{figure}

\begin{figure}
\vspace{-0.5cm}
$a)$ \\
\vspace{-0.5cm}
\begin{center}
    \includegraphics[width=0.975\textwidth,trim={0cm 0cm 0cm 0cm},clip]{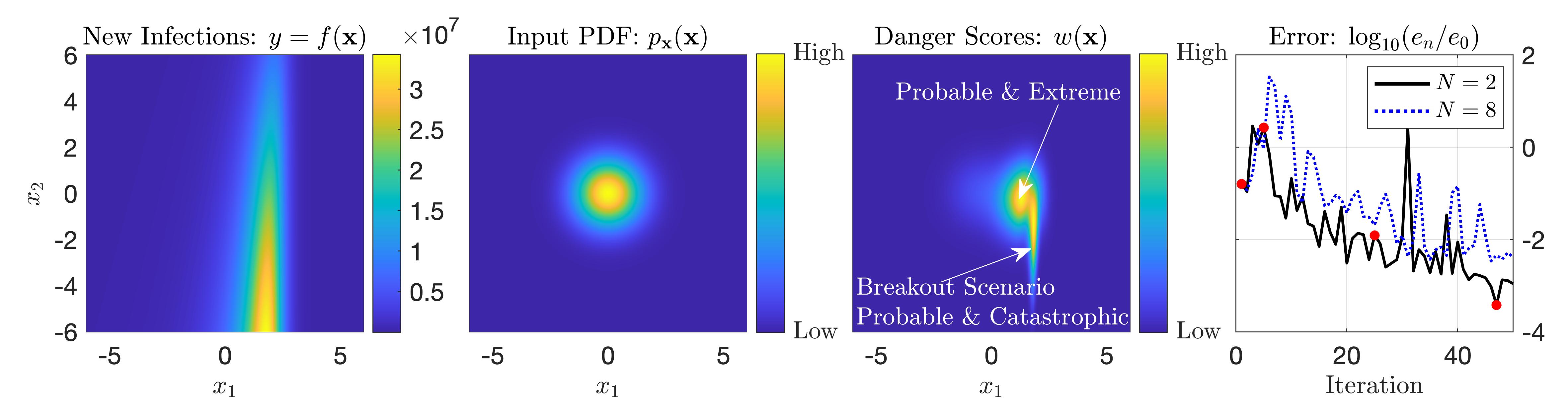}
\end{center}
\vspace{-0.75cm}
$b)$ \\
\includegraphics[width=1\textwidth,trim={0cm 0cm 0cm 0cm},clip]{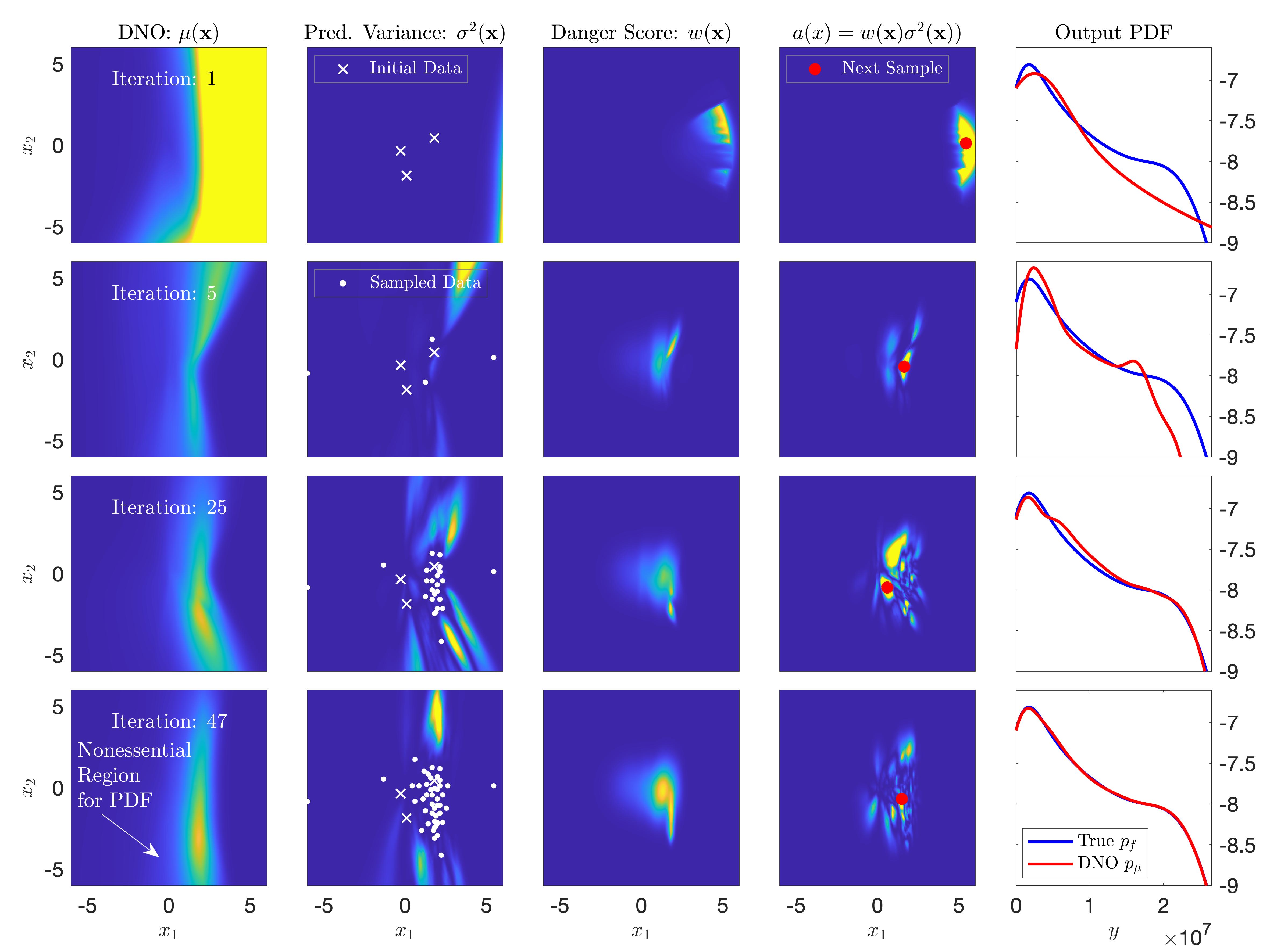}
\caption{\textbf{Nailing the Tail: Accurate PDF and danger score convergence in 50 samples.} $a)$ From left to right, the full deterministic response of new infections, $G(\mathbf{x})$, with respect to the two random parameters, $x_1$ and $x_2$, the probability distribution of the random parameters $x_1,x_2$, the underlying danger scores, $w(\mathbf{x})$ (with two regions of danger, \ethan{one probable and high magnitude (i.e. extreme) and one that is probable with catastrophically high magnitudes, constituting a breakout event}), and the $\log_{10}$ of the normalized ($e_0=10^7$) log-PDF error (equation \ref{eqn:PDF_error}), for the experiment performed in $b)$ using $N=2$. The red circles indicate the iterations shown in $b)$ as well as additional results for a case with $N=8$ ensembles. $b)$ One experiment of the 2D stochastic SIR model using three initial samples and iterated 50 times. The rows give the iteration number, 1, 5, 25, and 47 from top to bottom, respectively. The columns, from left to right, are the DNO approximation of the objective function, $\mu(\mathbf{x})$, given the training samples (initial + acquired), the samples acquired (where initial samples are white \textsc{x} and acquired samples are white circles) in the 2D parameter space and the predictive variance $\sigma^2(\mathbf{x})$, the calculated danger scores $w(\mathbf{x})$, the acquisition values $a(\mathbf{x})$ with the next acquisition sample denoted by a red circle, and the DNO approximated and true output PDFs. \ethan{Iteration 47 first column, identifies a high-magnitude infection region ignored by the algorithm, due to low input probability.} Animation links presenting 100 iterations for $N=2$ \href{https://drive.google.com/file/d/1xHpV8E03de_n606lhpwBOQm6GtLoinKk/view?usp=sharing}{(here)} and $N=8$ \href{https://drive.google.com/file/d/1kMZ1q2KXryYs3ttkbxNqIpXyukJe9FBj/view?usp=sharing}{(here)}.}
\label{fig:PandemicExample}
\end{figure}

\ethan{Our goal is to accurately quantify the probability distribution function (PDF) of a stochastic quantity of interest (QoI), $y$. The variable $y$ results from an observed random variable input, $\mathbf{x}$, that is transformed by the underlying system or map, $y = G(\mathbf{x})$. While the statistics of $y$ can be found through Monte Carlo sampling of the system, doing so is extremely inefficient. Instead, we aim to estimate an approximate map, $\tilde{G}$, via a surrogate model (e.g. Gaussian Process (GP) regression or DNO) trained on observed data pairs $\mathcal{D} = \{{ \mathbf{x}_i}, y_i \}_{i=1}^n$. The generated surrogate model may be sampled over $\mathbf{x}$ at substantially greater efficiency than the original system responsible for the observed  (training) data, $\mathcal{D}$. Using the Bayesian surrogate model we estimate the mean model, $\mu_n(\mathbf{x})$, by considering the mean over an $N-$ensemble of trained DNOs (using random initial weights). Note that for the case of GPs this mean can be calculated in closed form using standard expressions from GP regression.}


\ethan{Having the estimated mean-model for the QoI, $\mu_n(\mathbf{x})$, and the pdf of $\mathbf{x}$, $p_{\mathbf{x}} (\mathbf{x})$, we can estimate the pdf for the QoI, $p_{\mu_n}(y)$ via a weighted kernel density estimator (KDE). This is done by computing a large number of samples distributed over the input space with latin hypercube sampling, $\mathbf{x}_j$, evaluating them with the surrogate map, $\tilde{y}_j=\mu(\mathbf{x}_j)$, determining their probability of occurrence, $\alpha_j = p_\mathbf{x}(\mathbf{x}_j)$, and estimating the PDF, $p_{\mu_n}(y) = \mathrm{KDE}(\mathrm{data}=\tilde{y}_j, \mathrm{weights}=\alpha_j)$, using standard Gaussian KDE implementations. To emphasize the role of rare and extreme events, we assess the surrogate approximation by the error metric, 
\begin{equation}
 e_n = \int | \log_{10} p_{\mu_n} (y) - \log_{10} p_y (y)| \text{d} y, 
 \label{eqn:PDF_error}
\end{equation}
where $n$ is the algorithm iteration number (i.e. the number of training data points), while the integral is computed over a finite domain for the QoI, extended over the values that is interested to describe statistically. }

\ethan{Approximating the underlying map with a surrogate model may require substantial data depending on the complexity and dimension of the input space. To reduce the amount of necessary training data, we combine rare event statistics and Bayesian experimental design approaches to uncover the most critical data for training the surrogate model with surrogate map, $\tilde{G}$, that ultimately produces an accurate PDF, including the tails, of the QoI, $y$.}

Figure \ref{fig:Rogue_Pandemic} presents the key implications of our results, diagnosing the most dangerous future pandemic scenarios (left), \ethan{i.e. realizations of stochastic infection rates}, discovering seemingly benign waves that lead to dangerous rogue waves (center), \ethan{and identifying waves that lead to large structural stresses.} In each case, different scenarios are tied to a ``danger score'' or \textit{likelihood ratio}, as proposed by \cite{blanchard2021output},
\begin{equation}
    w(\mathbf{x}) = \frac{p_{\mathbf{x}}(\mathbf{x})}{p_\mu(\mu(\mathbf{x}))}.
\end{equation}
The likelihood ratio appropriately balances events that are probable and those that are extreme, hence it provides a danger score for any given event. As denoted for the pandemic model in figure \ref{fig:Rogue_Pandemic}, small danger scores are attributed to events that are either implausible or not extreme, while large danger scores relate to those that are both probable and extreme. However, any system's danger score requires knowledge of the true PDF, $p_{y}$, which is generally unknown and must be learned. Our approach efficiently learns this underlying distribution through dynamic application of the danger score with Bayesian experimental design and deep neural operators.

\subsection{Pinpointing Dangerous Pandemic Scenarios}


Figure \ref{fig:PandemicExample}~$a)$ and $b)$ \ethan{demonstrate how the proposed active learning framework leverages dynamically updating danger scores and predictive variances to efficiently sample the underlying system and learn the PDF of infections for a stochastic pandemic model.} The pandemic model is the simple Susceptible, Infected, Recovered (SIR) model, proposed by \cite{kermack1927contribution} and popularized by \cite{anderson1979population}, with a two-dimensional stochastic infection rate (see Appendix \ref{app:SIR}). We assess success as the log-PDF error, equation \ref{eqn:PDF_error}, in figure \ref{fig:PandemicExample}~$a)$ last column, between the true output distribution and the approximated distribution from the trained DNO, figure \ref{fig:PandemicExample}~$b)$ last column. See Section \ref{sec:danger_metrics} for details on computing the log-PDF error metric. 

Our framework quickly identifies the key regions of dynamical relevance and accurately recovers the significant properties of the underlying system, despite an initialization of only three data samples in the parameter space.  Using an ensemble of just two DNOs (see Appendix \ref{app:DON} for DNO implementation and section \ref{sec:DNO_UQ} for our application of ensemble methods), the algorithm iteratively provides an estimation of the underlying map (for computing $p_{\mu}$), predictive variance, $\sigma^2(\mathbf{x})$, and a danger score, $w(\mathbf{x})$. Together, the danger score and predictive variance create the likelihood-weighted uncertainty sampling (US-LW) acquisition function, $a(\mathbf{x}) =  w(\mathbf{x}) \sigma^2(\mathbf{x})$ \cite{blanchard2021output}, that identifies the sample within the parameter space with the greatest potential for learning the true output PDF. With the addition of each point, all fields dynamically change and bring the true and approximated output PDFs within greater agreement. By iteration 50, the danger scores have converged and the approximated output PDF is in error of less than $10^{-3}$. It is from this final danger score map that we derive the pandemic scenarios of figure \ref{fig:Rogue_Pandemic}. \ethan{This plot of danger scores includes two regions, one with probable and high-magnitude pandemic spike scenarios and another breakout scenario with catastrophic consequences (i.e. exceptionally high magnitude of infections). Each region is annotated in figure \ref{fig:PandemicExample}~$a)$ third column.} Additionally, figure \ref{fig:PandemicExample}~$a)$ last column, shows that increasing the ensemble size to $N=8$ provides little to no advantage. 

The critical aspect of this approach is the algorithm's reduction of a large parameter space to local regions of danger. Only regions that provide significant contributions to the output PDF are considered. Iteration 47, last row, first column, in figure \ref{fig:PandemicExample}~b) underscores this behavior. The algorithm has accurately reconstructed the output PDF, \ethan{yet by juxtaposing iteration 47 with figure \ref{fig:PandemicExample}~$a)$, we see that it has ignored a region where infections are of high magnitude located at $x_1 \approx 1.5$ and $x_2 = [-5,-6]$.} This region, and the remaining unexplored regions, provides negligible information about the QoI and is neglected by the acquisition function. This property is crucial for all systems where resources for experiments or simulation are limited or costly and it permits a significant reduction in training/acquired data as system complexity and dimensionality increases.

\subsection{Discovering and Predicting Rogue Waves}

\begin{figure}[t!]
\centering
$a)$  \hspace{2cm} $2D$ \hspace{2cm} $b)$\hspace{2.25cm} $4D$ \hspace{2.25cm} $c)$\hspace{2cm} $6D$ \phantom{PPPPPPPPPP}
\includegraphics[width=1\textwidth,trim={1cm 0cm 1.5cm 3cm},clip]{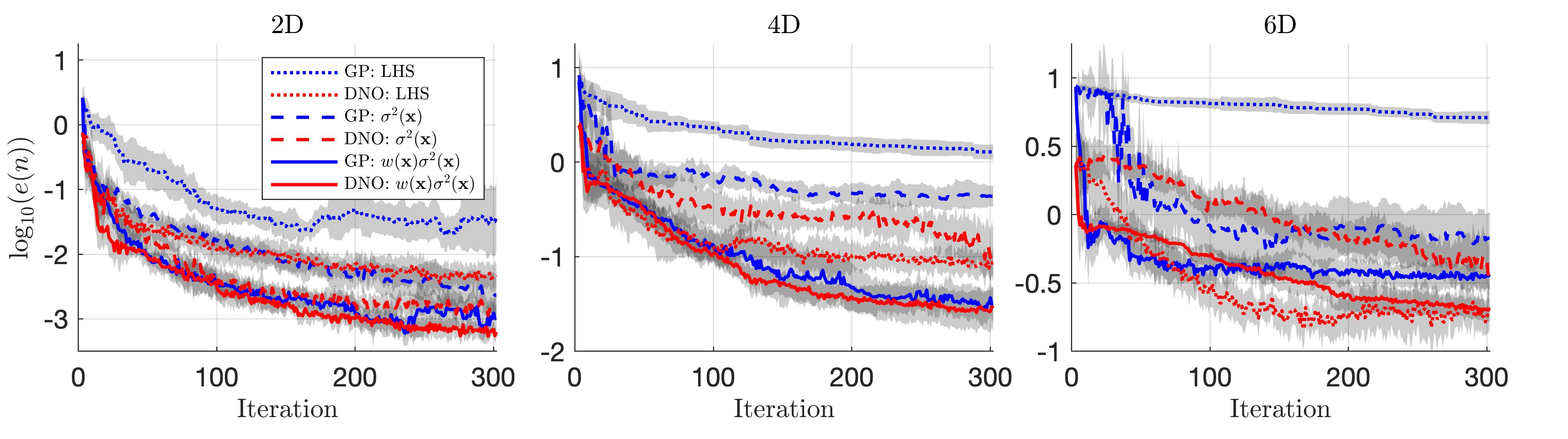} \\
$d)$ \hspace{1.25cm}  $8D$ Batching \hspace{1.25cm} $e)$ \hspace{0.75cm}  $8D$ Ensemble Sizes, $N$ \hspace{0.75cm}  $f)$ \hspace{0.95cm}  $8D$ Variance of $N$ \phantom{PPP} \\
\includegraphics[width=0.325\textwidth,trim={0.3cm 0cm 1.5cm 0cm},clip]{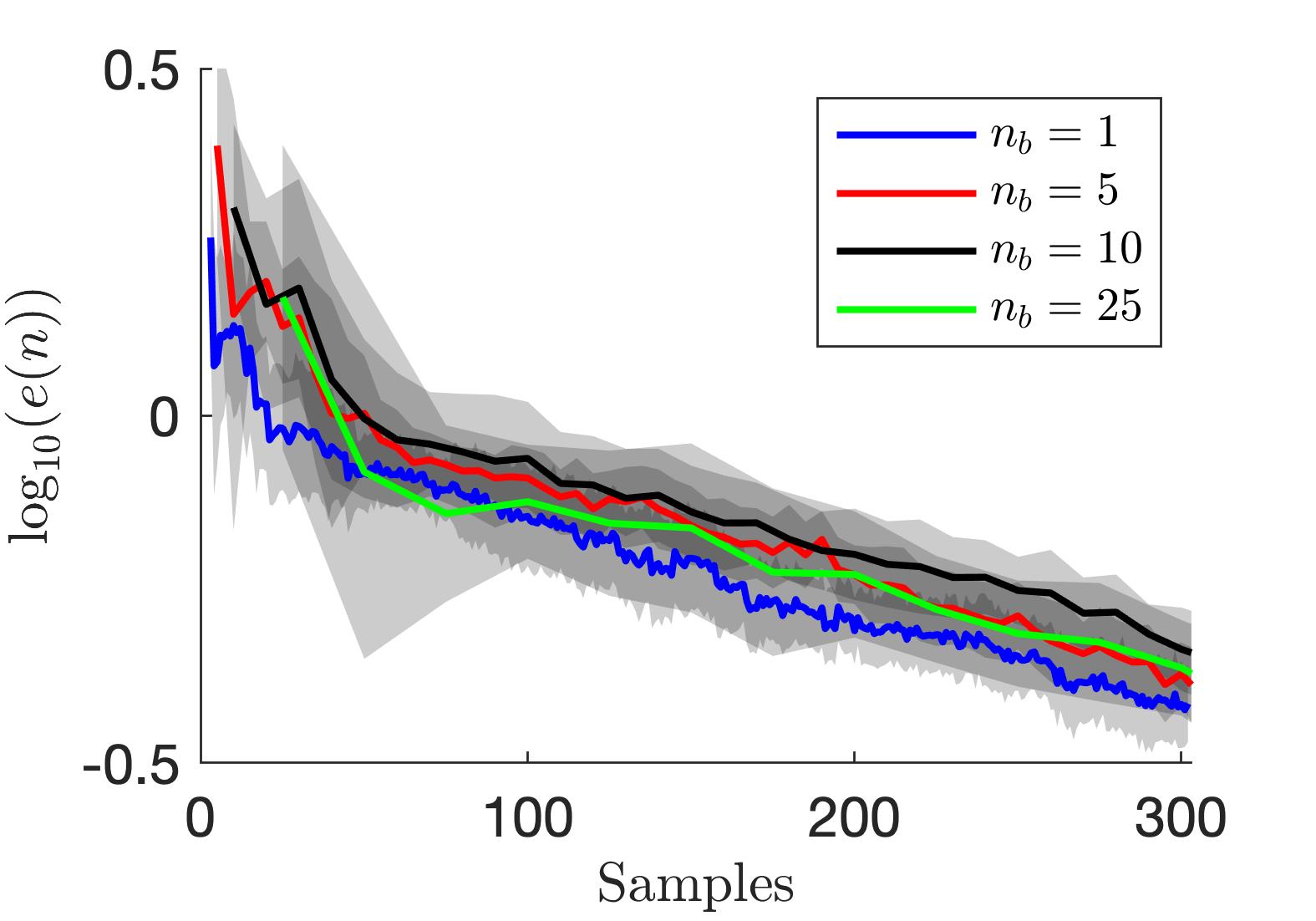}
\includegraphics[width=0.325\textwidth,trim={0.3cm 0cm 2cm 0cm},clip]{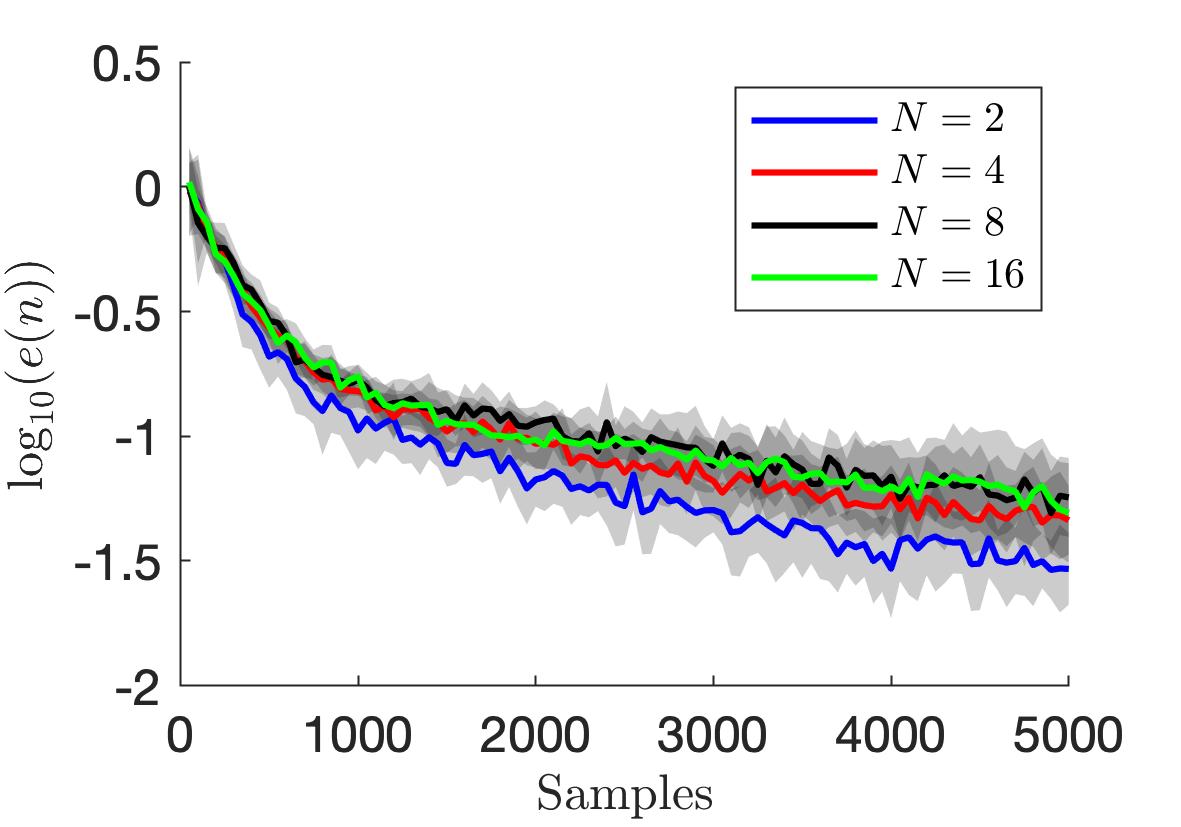}
\includegraphics[width=0.325\textwidth,trim={0.3cm 0cm 2cm 0cm},clip]{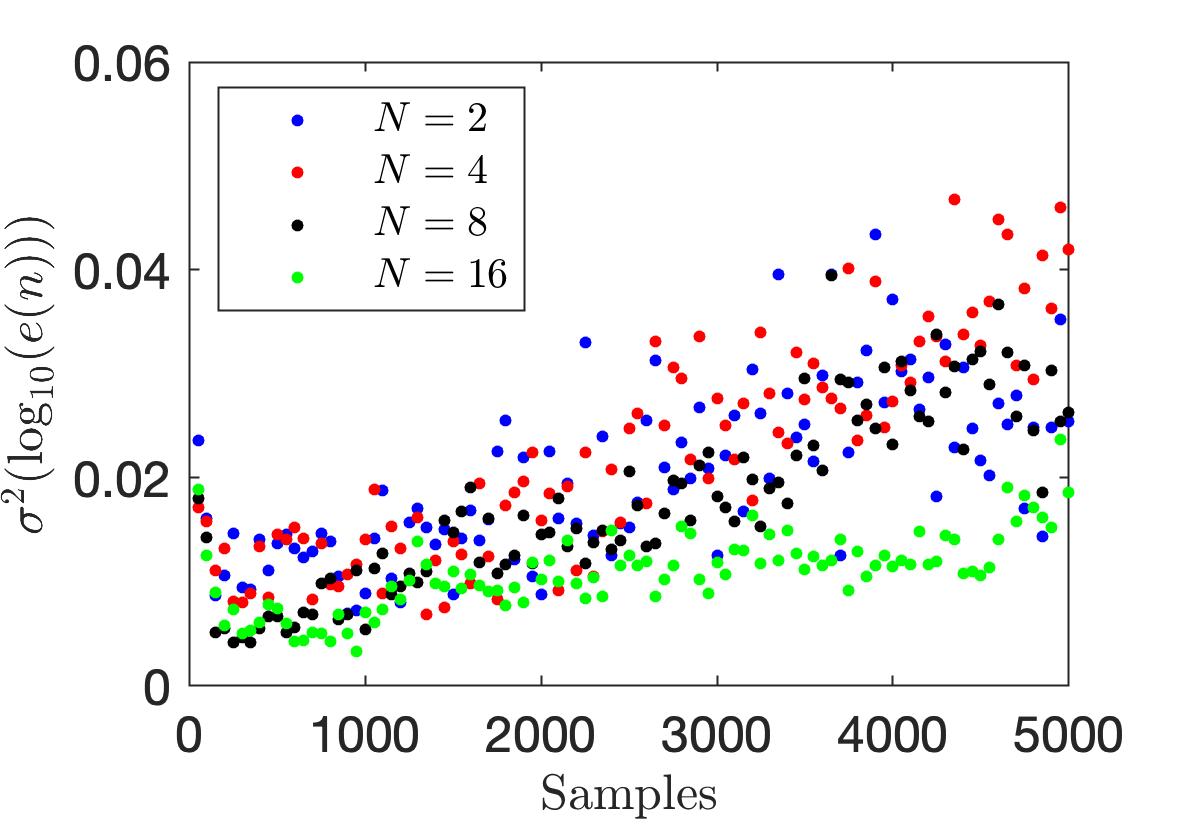} 
\caption{ \textbf{Accelerated convergence with DNOs plus extreme acquisition functions regardless of dimensionality.} $a)-c)$ Median log-PDF error, equation \ref{eqn:PDF_error}, of 10 independent and randomly initialized experiments, where shaded regions denote one standard deviation for each case (i.e. $\pm \sigma(\epsilon)$), considering three acquisition functions and GP and DNO surrogate models. $a)-c)$ are for $2D$, $4D$ and $6D$ initial conditions (ICs), while $d)-f)$ represent results from $8D$ ICs. \textbf{Parallel acquisition and shallow ensembles bring computational efficiency without performance loss.} $d)$ provides errors related to different batch sizes, $n_b$, per iteration, $e)$ gives performance differences relative to the DNO ensemble size (100 iterations at batch size $n_b=50$), and $f)$ presents the \ethan{variance of the log-PDF error of show in $e)$}.}
\label{fig:1d_3D_Errors}
\end{figure}

\ethan{We now train a surrogate model for rogue wave prediction by actively discovering the probable initial conditions (ICs), or precursors, responsible for such phenomena. Here we present a proof of concept with a dispersive nonlinear wave model proposed by Majda, McLaughlin, and Tabak (MMT) \cite{majda1997one,cai1999spectral} for 1D wave turbulence. The same model has also been used as a prototype system to model rogue waves \cite{zakharov2001wave,zakharov2004one,pushkarev2013quasibreathers, cousins2014quantification} (Appendix \ref{app:NLS}). }

\ethan{We seek to map initially observed waves, $u(x,t=0)$, where $x$ is the spatial variable, to the QoI: the future spatial maximum $G(\mathbf{x}) = || Re(u(x,t=T;\mathbf{x}))||_{\infty}$, where $T$ is a prescribed prediction horizon. We note that MMT has complex solutions and therefore the ICs are also complex valued.  In a real application, other quantities, such as the short time derivative, would accompany the initial condition, instead of an imaginary component, to provide wave speed.} We now investigate this complex and highly nonlinear problem by systematically scaling the dimensionality and expanding to larger datasets, where GPs begin to fail.




The DNO-BED framework, using the likelihood-weighted uncertainty sampling acquisition function ($a(\mathbf{x}) = w(\mathbf{x}) \sigma^2(\mathbf{x})$), efficiently minimizes the error between the approximated and true PDF of the QoI when compared to GPs (detailed in Appendix \ref{app:GP}) or other common BED sampling strategies, such as uncertainty sampling (US, $a(\mathbf{x}) = \sigma^2(\mathbf{x})$) and LHS, as shown in figure \ref{fig:1d_3D_Errors}~a)-c).
This is highlighted as dimensionality (i.e., complexity, details on dimensionality provided in Appendix \ref{app:NLS}) \ethan{of the parameterized ICs} is increased from $2D$ to $6D$, where GPs begin to break down. 

This framework brings favorable results for reducing computational cost in high dimensions as both batching samples via parallel selection of multiple acquisition points, figure \ref{fig:1d_3D_Errors}~$d)$, and the use of shallow ensembles of only $N=2$ members, figure \ref{fig:1d_3D_Errors}~$e)$, perform without loss.
Computed at $8D$, figure \ref{fig:1d_3D_Errors}~$d)$ shows that regardless of choosing $n_b=1,5,10$, or even 25 samples per iteration, over 300 samples, does not result in a loss of performance with our framework. Further, we describe in Appendix \ref{app:monte} that the batched samples come from several regions of local optima and the use of Monte Carlo methods is substantially more efficient for identifying these acquisition samples than standard \textsc{Python} optimizers. See Section \ref{sec:batch} for details on batching implementation. These observations are a critical result for scaling our framework, as more complex systems inevitably will require more data. 

Unexpectedly, \textit{not only do just two ensemble members, $N=2$, perform well, they consistently outperform larger ensembles,} $N>2$, in figure \ref{fig:1d_3D_Errors}~$e)$ (\ethan{where 100 iterations of batch size $n_b=50$ are applied to the $8D$ case.).} This result appears to disagree with the natural hypothesis that a larger set of ensembles would provide uncertainty estimates with greater fidelity, leading to better performance of our sequential search methods. Clearly, the latter is not the case from our results, yet neither can it be that $N=2$ ensembles provide a predictive variance of greater fidelity than $N=16$. Figure \ref{fig:1d_3D_Errors} $f)$ permits both concepts to be true. It shows that the greater the ensemble size, the smaller the variance between the error trajectories of independent experiments. This observation agrees with the idea that larger ensembles lead to a higher-fidelity predictive variance, but that greater fidelity leads to \textit{consistency} rather than \textit{performance} for this sequential search technique. We believe using small $N$ imposes a greedy search, in a similar fashion to Thompson sampling \cite{chapelle2011empirical}. Regardless, the consistent observation that $N=2$ is not only viable, but perhaps preferable, has massive implications for minimizing computational costs for ensemble approaches. 

\subsubsection{Rogue Wave Discovery in $20D$}

\begin{figure}
\centering
\phantom{} $a)$ \hspace{2.5cm} $20D$ \hspace{2.7cm} $b)$ \hspace{2.5cm} $20D$ \phantom{PPPPPPPPP} \\
\phantom{} \hspace{0cm} \includegraphics[width=0.375\textwidth,trim={0cm 0cm 0.5cm 3cm},clip]{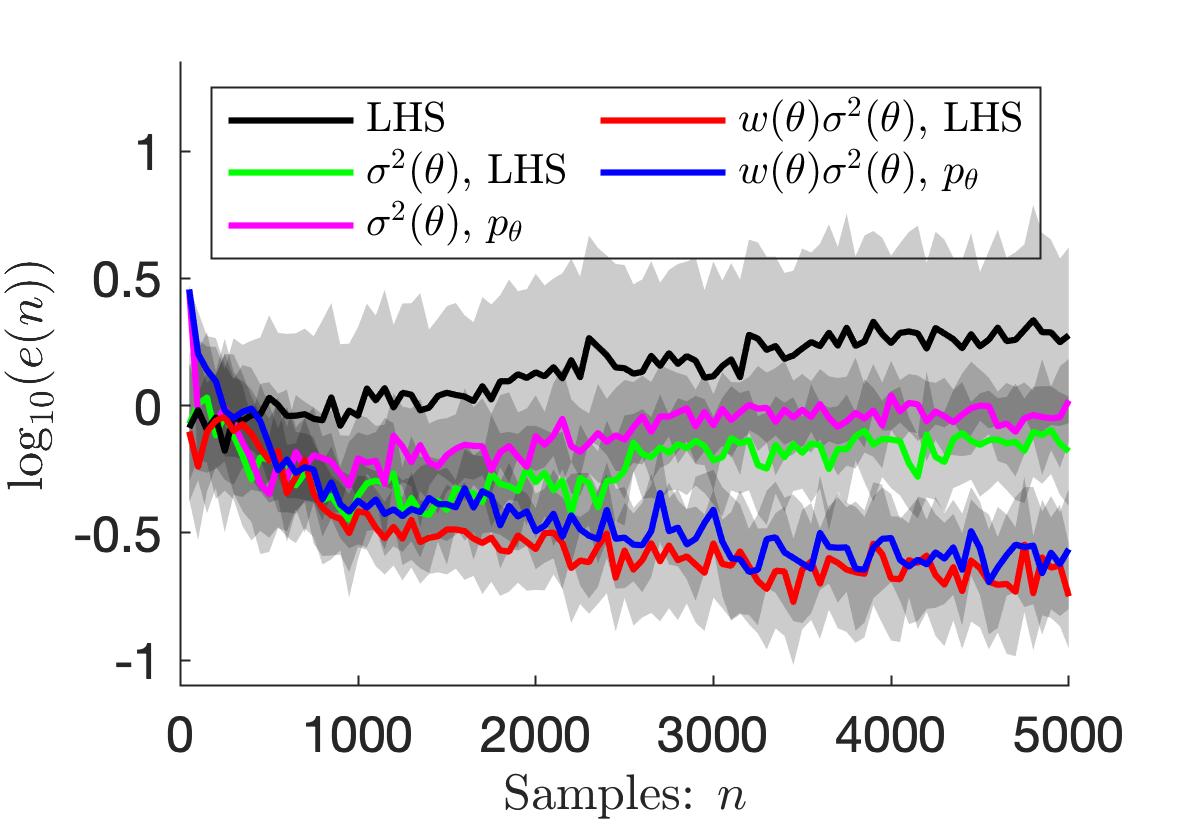} \includegraphics[width=0.375\textwidth,trim={0cm 0cm 0.5cm 3cm},clip]{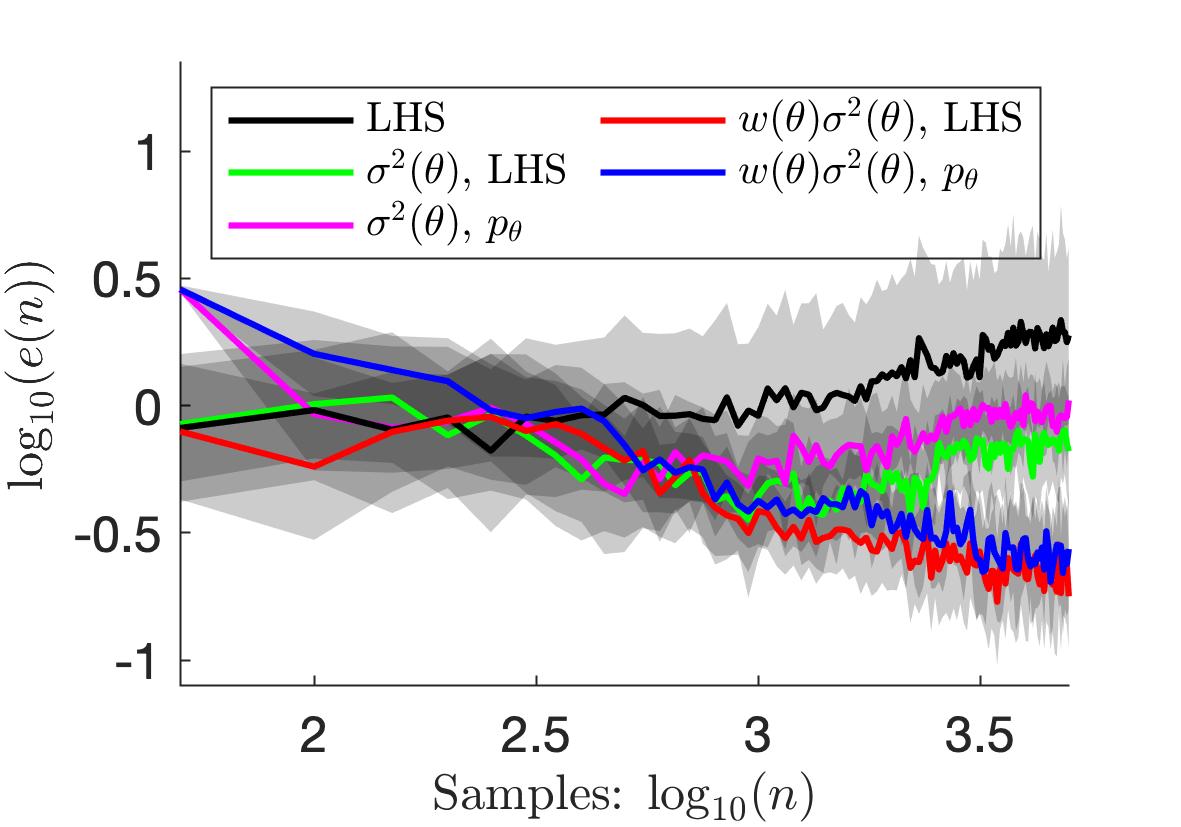}\\
$c)$ \hspace{0.15cm} \small{10 Initial Samples \textit{with} Extremes (LHS)\phantom{PP} 10 BED Acquired Samples (LHS)} \phantom{PPPPP} \\
\includegraphics[width=0.75\textwidth,trim={0cm 0cm 0.5cm 3cm},clip]{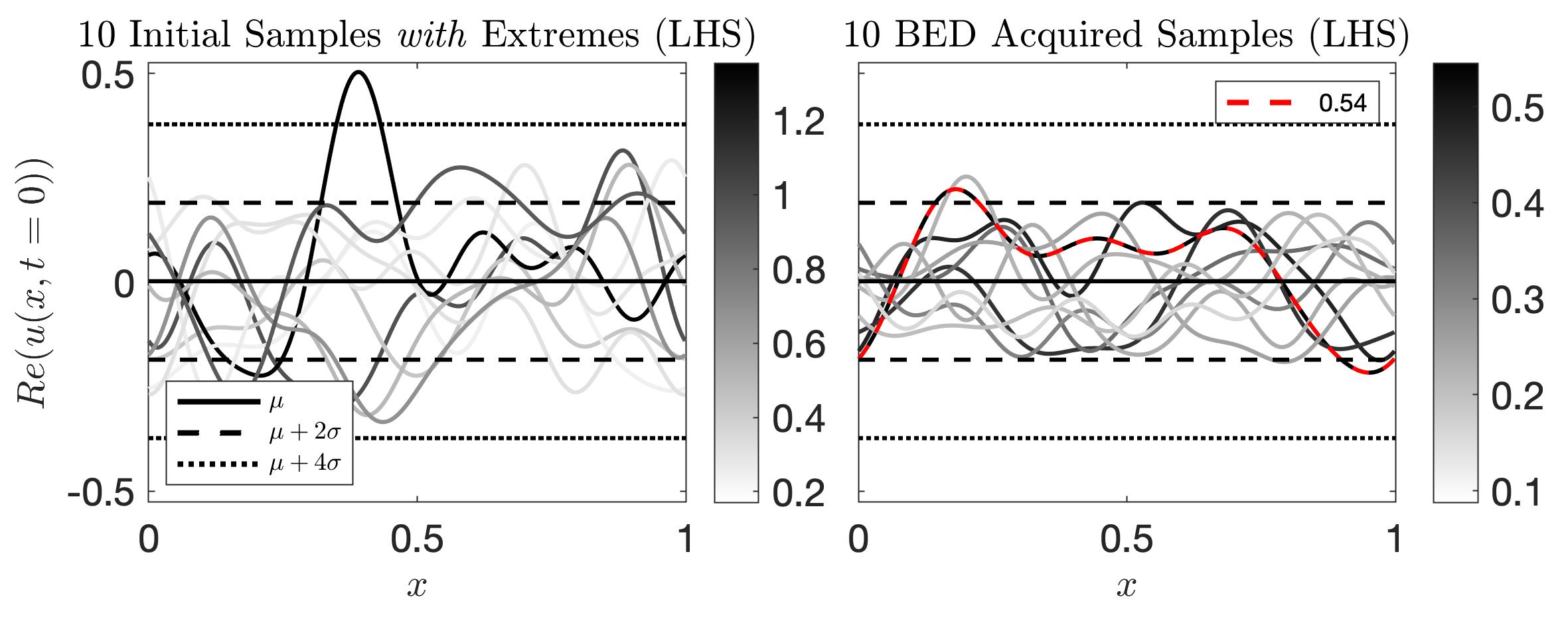}\\
$d)$ \hspace{0.05cm} \small{10 Initial Samples \textit{without} Extremes ($p_{\mathbf{x}}$)\phantom{PP} 10 BED Acquired Samples ($p_{\mathbf{x}})$} \phantom{PPP} \\
\includegraphics[width=0.75\textwidth,trim={0cm 0cm 0.5cm 3cm},clip]{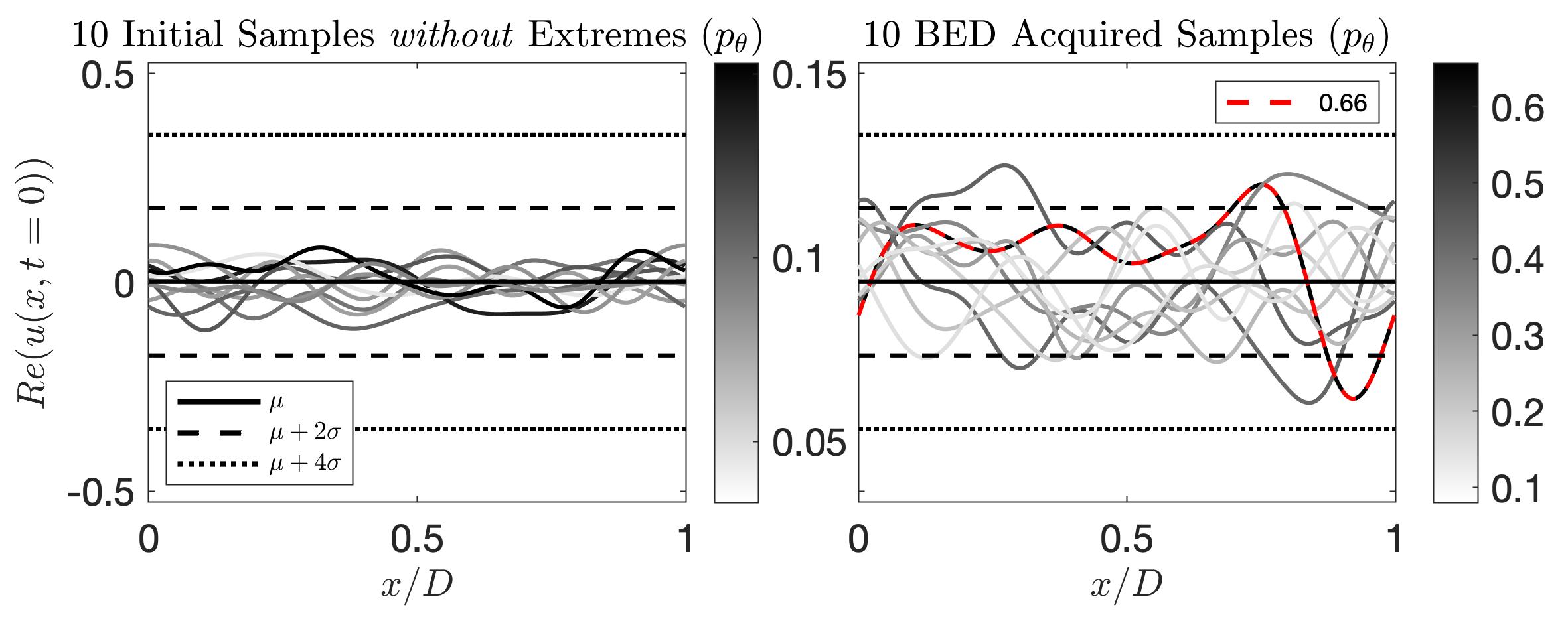}
\caption{\textbf{Robustness to dimensionality and initial data, i.e., with or without extremes.} $a)$ The log-PDF errors for three acquisition functions (LHS, US, and US-LW) for the 20$D$ problem, with initial samples sampled via LHS and the prior $p_x$ for US and US-LW. $b)$ provides the same results in log-log form. $c)$ gives the real value of 10 complex LHS initial input functions $u(x, t=0)$, colored with their corresponding QoI, $||Re(u(x,t=T))||_{\infty}$, (left) and 10 DNO-BED acquired functions, via US-LW at iteration 97 (right). $d)$ gives 10 initial input functions sampled from the prior distribution $p_{x}$ (left) and the 10 acquired function at iteration 74 (right).}
\label{fig:Lines_LHS}
\end{figure}

Equipped with the computational advantages of $N=2$ ensemble members and large batch sizes ($n_b=50$), figure \ref{fig:Lines_LHS}~$a)-b)$ shows that even at $20D$ our approach can recover the QoI PDF. The other acquisition functions not only perform poorly, but perform worse as more data is acquired. This observeration appears to be related to the phenomenon known as \ethan{sample-wise} ``double descent’’, and many researchers have observed this behavior throughout machine learning procedures, from classification to regression problems \cite{nakkiran2021deep}.  \ethan{Sample-wise} double descent is associated with instabilities in the surrogate model, a product of over-fitting. More data and greater complexity results in an over-parameterization of the provided data. Temporarily, this leads to inferior generalization until providing the surrogate model with sufficiently larger datasets. 

\textit{The proposed acquisition function clearly avoids \ethan{log-PDF error} double descent \ethan{for the $20D$ problem}}. \ethan{While we do not explicitly detail the mechanism behind this observation here, we refer to a parallel study \cite{pickering2022fomo} on this exact problem in $8D$ (showing elimination of both mean square error and log-PDF error double descent, as well as other examples using GP surrogate models) and briefly outline why the acquisition brings this beneficial behavior. As discussed in \cite{pickering2022fomo}, double-descent is eliminated by only selecting data that critically contribute to the observed dynamics of the system.} Unlike US-LW, the data chosen by LHS and US methods are not inherently important to recovering the true PDF and therefore induce misleading complexity to the underlying regression task. This observation further underscores the value of our acquisition function, as it  systematically prevents over-fitting and unwarranted model complexity.



Additionally, we find that regardless of the origin of initial samples, containing extremes (chosen by LHS) or without extremes (from the prior, $p_\mathbf{x}$), the method achieves similar error metrics in figure \ref{fig:Lines_LHS}~$a)-b)$.
\ethan{To elucidate the physical meaning of these results, figure \ref{fig:Lines_LHS} $c)$ and $d)$ present the physical manifestations of the $20D$ initial conditions, i.e. the real component of $u(x, t=0)$, that include extremes and those that do not, respectively. The horizontal lines denoting the mean and standard deviations, $\mu, \sigma$, refer to the statistics of the normally distributed ICs, $u(x, t=0)$, \textit{not} the QoI. Instead, the QoI is observed \textit{after evolving} the presented ICs by $T$ and the associated QoI value is represented by the color of the line. Focusing on figure \ref{fig:Lines_LHS}~$c)$ (left), several of the initial LHS ICs are already extreme, as portions of the ICs already exceed 3 standard deviations (the common delimiter of extremes for normal distributions), while figure \ref{fig:Lines_LHS}~$d)$ (left) gives ICs samples from the prior and are nearly bounded by one standard deviation. Despite these clearly different sets of initial ICs, the algorithm samples similar ICs in the right plots of figure \ref{fig:Lines_LHS}~$c)$ and $d)$, and achieves similar error metrics in figure \ref{fig:Lines_LHS} $a)-b)$. These ICs are approximately bounded within $\pm 2\sigma$, meaning they are neither extreme (yet) nor common, but sit on the periphery of a dynamical instability that may lead to an extreme event. Therefore, the method is able to uncover the seemingly benign conditions that lead to extremes, regardless of the initial training set. Finally, with respect to the IC statistics, observed QoIs above $\approx 0.3$ constitute an extreme event at $t=T$.} 

\vspace{-0.75cm}
\ethan{\subsection{Efficient Estimation of Structural Fatigue for Ship Design}}
\label{sec:lamp}

\ethan{Uncovering the statistical signature of a marine vessel's Vertical Bending Moment (VBM) at the midship section is critical to estimating fatigue lifetimes. Caused by cyclic hydrostatic pressure forces and the slamming of the ships bow into oncoming waves \cite{sapsis20b, sapsis21b, belenky21}, large VBMs increase the potential for microfracture nucleation and propagation \cite{serebrinsky05, khan07, chasperis10}. Just as in our previous examples, these forces are stochastic processes that result in unique VBM statistics with each unique ship design (e.g. the Office of Naval Research topsides flare variant studied here) via an underlying nonlinear operator. Performing either tow-tank or numerical experiments is expensive and time consuming, limiting the potential for optimal structural design. Here we apply our method to a proprietary code, LAMP (Large Amplitude Motion Program, v4.0.9, May 2019) \cite{lin10}, in order to calculate the \textit{expensive} forward problem of a specific VBM response to a specific wave episode for informing ship design.}

\begin{figure}
$a)$ \hspace{7cm} $b)$ \\
\includegraphics[width=0.45\textwidth,trim={0cm 0cm 0cm 0cm},clip]{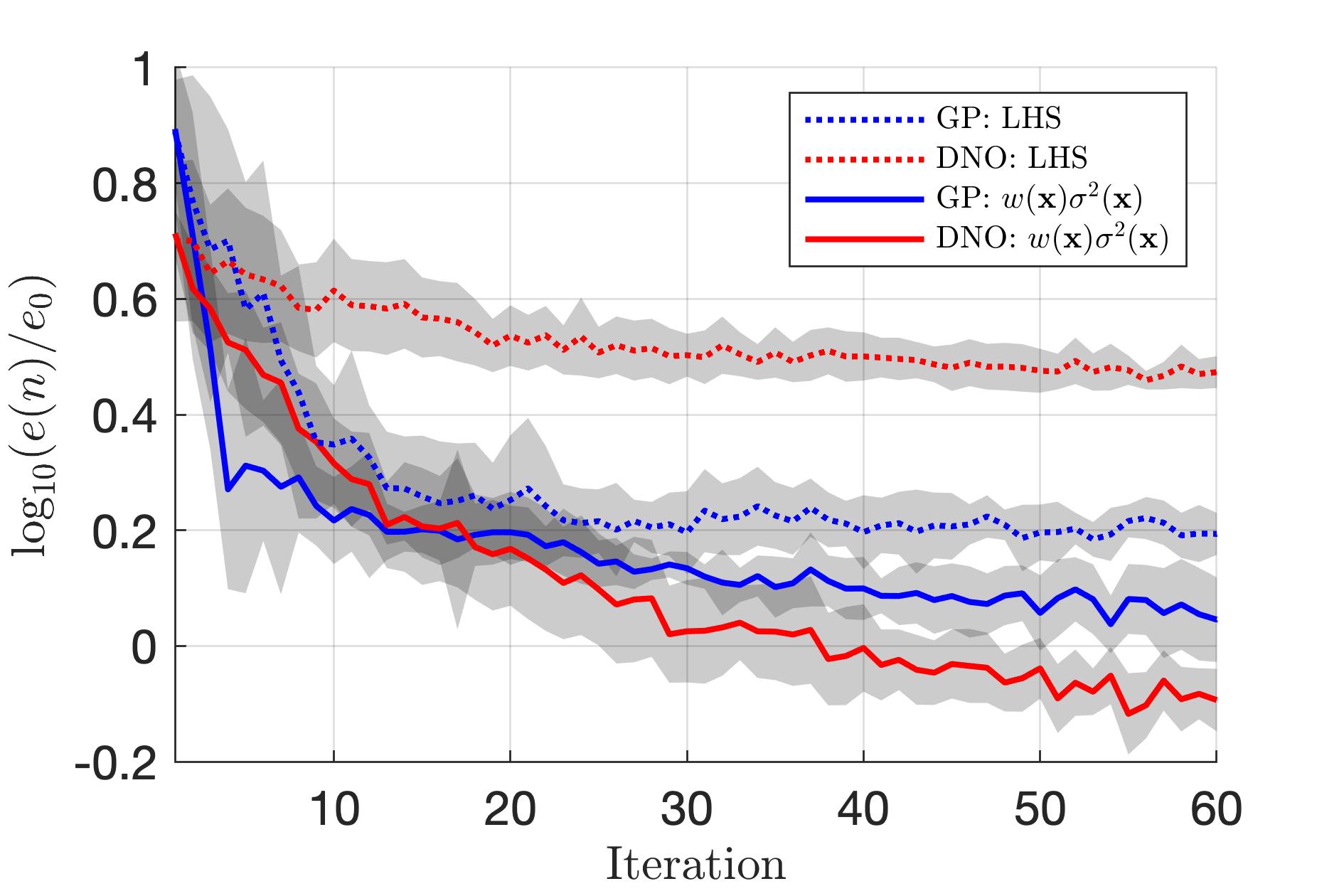}
\includegraphics[width=0.45\textwidth,trim={0cm 0cm 0cm 0cm},clip]{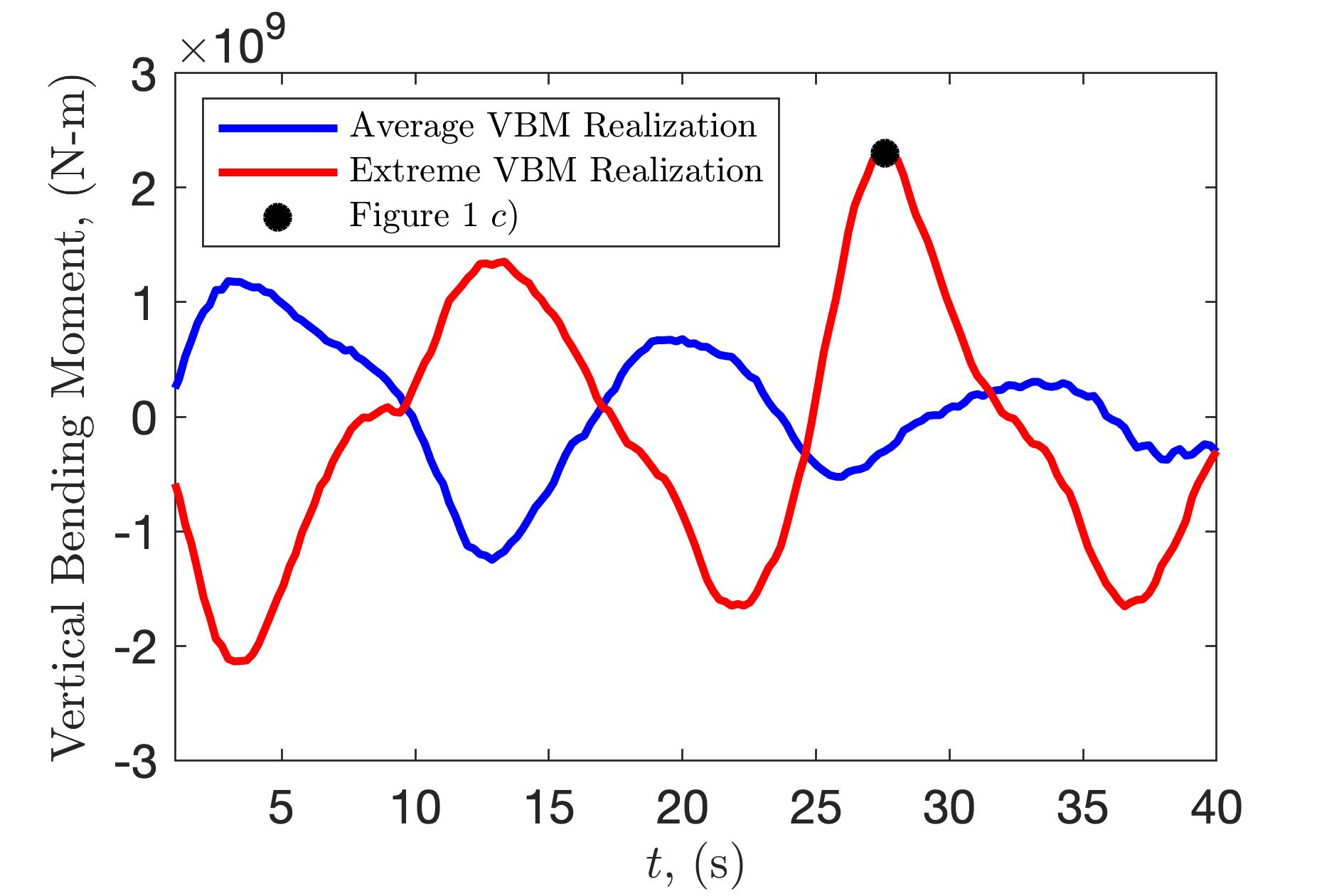}
\caption{\textbf{a) Efficient learning of fatigue statistics for ship design.} Median normalized ($e_0=5\times 10^8$) log-PDF errors for 10 experiments of VBM statistics for GP and DNO using LHS and US-LW, where the shaded regions denote one standard deviation (i.e. $\pm \sigma(\epsilon)$). \textbf{b) Time series of average (top) and extreme (bottom) VBM stresses from LAMP.} Two representative VBM curves of average and extreme loads on the ship from stochastic ocean waves. The marker reflects the instantaneous VBM sustained by the ship in figure \ref{fig:Rogue_Pandemic}.}
\label{fig:LAMP_Learning_Curves}
\end{figure}

\ethan{Applying the framework to a $10D$ subspace of finite-time ocean wave episodes impacting the ship modeled in LAMP, we again find US-LW DNO is superior in learning the VBM (QoI) statistics compared to other methods in figure \ref{fig:LAMP_Learning_Curves}$~a)$.
To highlight the relative ease of implementation of the DNO based approach, we emphasize that GPs were very carefully tuned (optimization of hyperparameters using additional simulations that took significant time and effort) compared to minimal tuning efforts for DNOs. Additionally, figure  \ref{fig:LAMP_Learning_Curves}$~b)$ presents two representative realizations, one average and one extreme, of the VBM time series felt by the ship through a set of ocean waves. The highlighted marker on the extreme realization at $t=27$ represents the instantaneous VBM load that the ship encounters during an instant presented in the introduction, figure \ref{fig:Rogue_Pandemic}$~c)$.}

\section{Discussion} \label{sec:conclusions}

\begin{figure}
$a)$ \\
\includegraphics[width=1\textwidth,trim={0cm 0cm 0cm 0cm},clip]{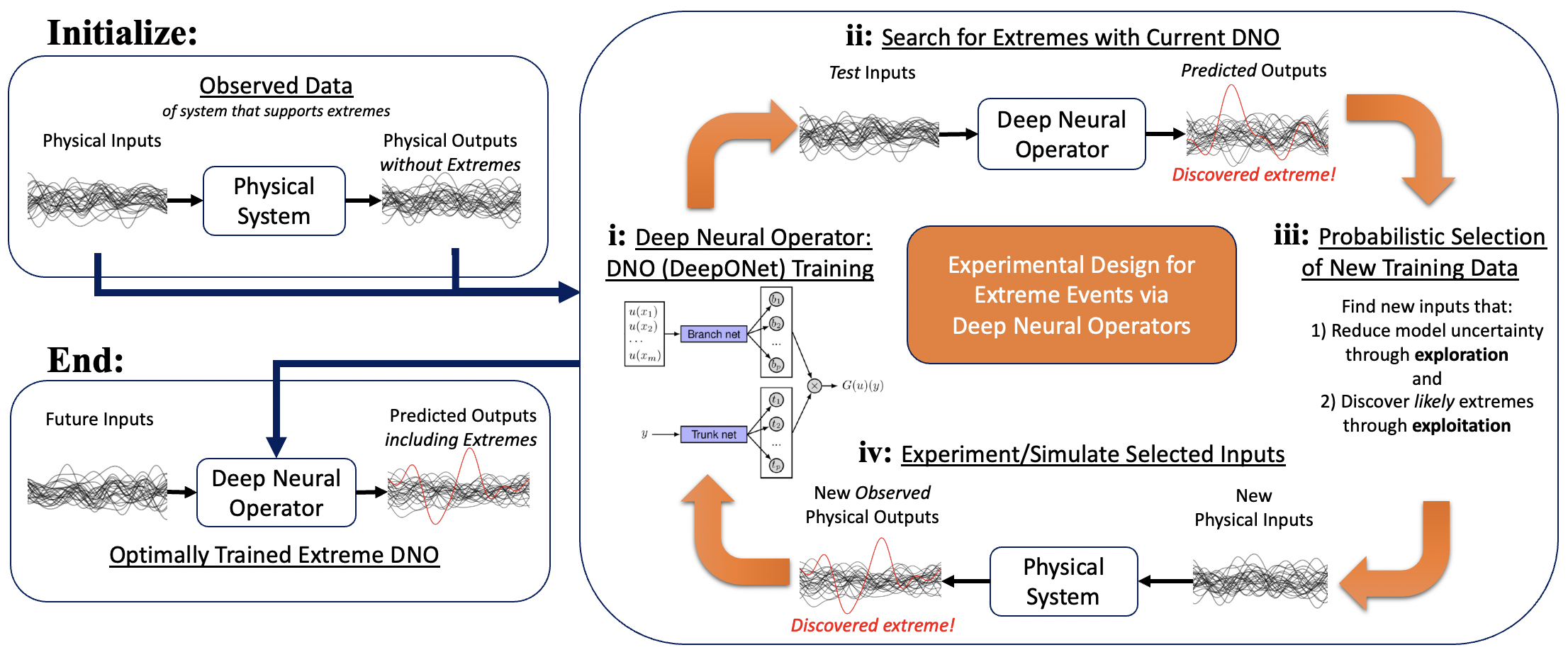} \\
$b)$ \\
\includegraphics[width=1\textwidth,trim={0cm 0cm 0cm 0cm},clip]{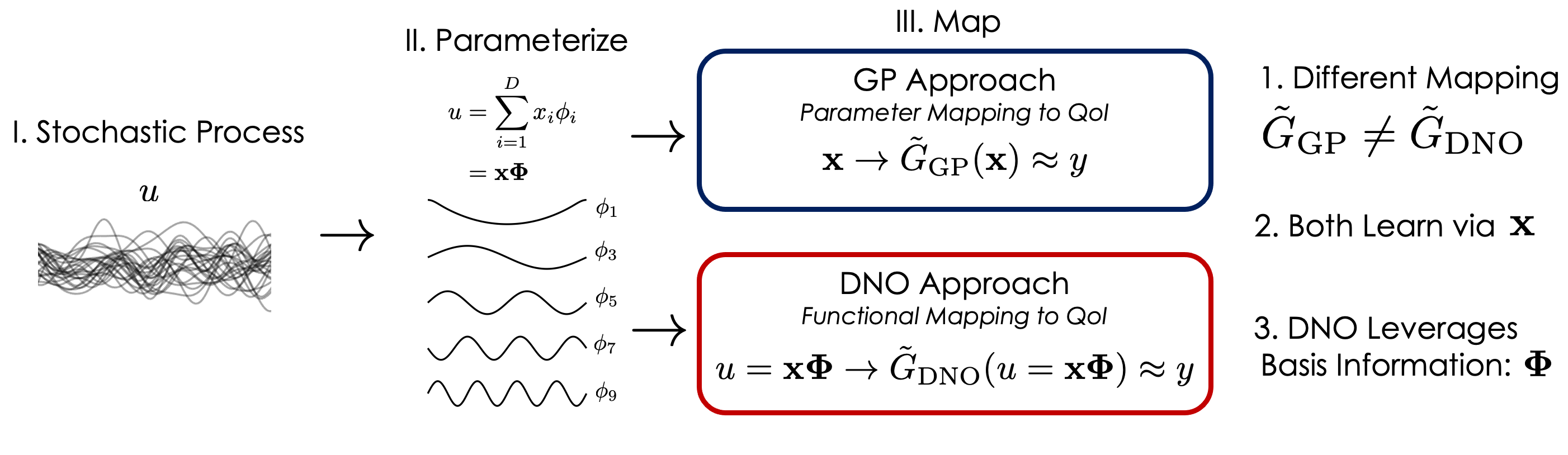}
\caption{\textbf{a) Efficient and robust DNO+BED framework for discovering and quantifying extremes.} An overview of the proposed Bayesian experimental design framework with deep neural operators and novel acquisition functions for discovering extremes. \textit{Initialize)} with a set of observed physical input-output pairs, retained in their functional form. \textit{i)} Pass functions to an ensemble of DNOs to learn sparse representations of the underlying system. \textit{ii)} Perform a fast Monte Carlo search of the DNO functional space for extremes. \textit{iii)} Compute statistics over a Monte Carlo ensemble and select new input functions that both explore and exploit the space for extremes. \textit{iv)} Evaluate proposed inputs on the underlying experiment or simulation, record outputs (QoIs), and pass to \textit{i)}. Repeat \textit{i)-iv)} until statistics are converged or resources are depleted. \textit{End)} with an optimally trained DNO that supports prediction of extreme events. \ethan{\textbf{b) DNO leverage functional information for mapping to the QoI.} Both GPs and DNOs learn a stochastic process via a paramaterization of a stochastic process, but differ in that GPs map the parameter space to the QoI while DNOs map the functional realization of the parameter space to the QoI.}}
\label{fig:ExpDesignSchematicDisc}
\end{figure}


The discovery, characterization, and ultimate prediction of extreme events remains a grand challenge across application domains. Presented in figure \ref{fig:ExpDesignSchematicDisc}, we propose an AI-assisted, model-agnostic infrastructure that paves the way for efficient and robust inference of these critically important events. By combining deep neural operators, Bayesian experimental design, and tailored acquisition functions, we create a framework that can accurately discover, characterize, and predict extreme events in high-dimensional and \ethan{stochastic} nonlinear systems ranging from societal to physical. We demonstrate the efficacy of this framework here by efficiently discovering and predicting rogue waves, reducing the number of samples from $10^7$ to $10^2$, accurately pinpointing the most dangerous pandemic spikes, and efficiently quantifying ship stress statistics for informing reliable ship design. 

This approach is not only superior to other techniques, namely Gaussian Processes (GPs), but comes with several expected and unexpected, yet preferable, consequences. As expected, we show that the use of DNOs brings both greater generalization accuracy and a propensity for extremely large datasets. Unexpectedly, we find that shallow ensembles of only \textit{two} ensemble members perform remarkably well and that the use of batches of optimal sample samples acquired in parallel, compared to step-by-step, or serial, acquisition, does not hinder BED performance. We also find that Monte Carlo approaches for selecting these optimal samples perform better than standard optimizers, which routinely fall victim to the highly non-convex nature of high-dimensional optimization. These results bring crucial real-world benefits to our approach. Batching permits parallel acquisition of data, while small ensembles keep computational costs tractable. 

Our method provides accurate extreme event quantification \textit{regardless} of the state of initial data and eliminates sample-wise ``double descent'' in the $20D$ example. While we refer to a parallel study, \cite{pickering2022fomo}, for an in-depth analysis on how the method eliminates double-descent, both observations underscore the importance of sampling the \textit{most informative data rather than more data}. Double descent is avoided as the acquisition function only selects samples that significantly contribute to the observed dynamics of the system, while ignoring data that would bring added and unnecessary complexity. On the other hand, the method is robust to the initial data provided and quickly directs attention to the most informative samples whether or not extremes have been observed. 



\ethan{Despite these advantages and the generality of the approach to any stochastic nonlinear system with rare events, we caution readers that our results are empirical and do not come with robust guarantees. This is compounded by the non-analytical nature of neural networks, which limit hope that guarantees may ever be found. Thus, implementation must be performed carefully and systematically. Training the DNOs is not trivial and requires finesse, while the acquisition function, although not observed here, could fall prey to unforeseen pathological cases. To help readers implement the method, Appendix \ref{app:Tips} provide several tips as well as general implementation concerns that we either experienced or recognized during implementation.}


The framework also provides modularity for the chosen surrogate models and acquisition functions, so long as the parameter and function spaces are appropriately defined. Critically, we separate the two, \ethan{as shown in figure \ref{fig:ExpDesignSchematicDisc}~$b)$. In GP implementations, the parameter space of a stochastic process is used for both regression and searching. Instead, our implementation of DNOs perform regression in the functional space, leveraging the typically disregarded basis functions associated with the paramaterization, while the search algorithm is performed in the parameter space via a forward DNO coupling with the associated basis functions. It is this maintenance of the functional representation that provides improved generalization across the parameter space.} 

As such, any arbitrary neural architecture leveraging this distinction may be implemented, such as  standard feed-forward neural networks (NNs), Fourier Neural Operators \cite{li2020fourier}, Convolutional NNs, Recurrent NNs, Long-Short Term Memory, among others. In fact, this work did not explicitly investigate the full value of DeepONet for operator learning for BED. Instead, only the standard feed-forward branch NN was used. Using the complete operator machinery only requires adjusting the parameters related to the operator and trunk. Here, we kept these parameters constant. Finally, while we focus on Bayesian experimental design, slight adjustments to the choice of acquisition function allow for Bayesian Optimization tasks \cite{yang2021output} or for approaching other metrics of interest (e.g. mean squared error).
 
In conclusion, we believe we have demonstrated an equation-agnostic framework that: (1) efficiently discovers extreme events, (2) is computationally tractable, and (3) presents a straightforward implementation on any stochastic input-output system. This creates a unique opportunity for experimentalists and computationalists alike to investigate and quantify their stochastic systems with respect to extreme behavior, whether that behavior originates from societal or physical systems or has beneficial or catastrophic consequences.

\section{Methods} \label{sec:Methods}


Our approach to Bayesian experimental design is detailed in figure \ref{fig:ExpDesignSchematicDisc} and consists of two critical components, \textit{data selection criteria} and the \textit{surrogate model}. Algorithm~\ref{algorithm1} formalizes the iterative steps taken in figure \ref{fig:ExpDesignSchematicDisc} for efficiently training a surrogate model with minimal data selection.

\begin{algorithm}
  \caption{Sequential search for active learning/training of DNOs and GPs .}\label{algorithm1}
  \begin{algorithmic}[1]
    \STATE \textbf{Input:} Number of iterations: $n_{iter}$
    
    \STATE \textbf{Initialize:} Train GP/DNO on initial dataset of input-output pairs \hspace{2cm} \phantom{PPPPPPPPPPPPPP} \hspace{2cm} GP:\hspace{0.25cm } $\mathcal{D}_{0}=\left\{\mathbf{x}_{i}, y_i=G(\mathbf{x}_{i}) \right\}_{i=1}^{n_{\text {init }}}$ \hspace{1cm} DNO: \hspace{0.25cm }$\mathcal{D}_{0}=\left\{\mathbf{x}_{i}, y_i=G(\mathbf{x}_{u,i} \mathbf{\Phi}(x_1, ... x_m),\mathbf{x}_{z,i}) \right\}_{i=1}^{n_{\text {init }}}$

    \STATE \textbf{for} $n=1$ \textbf{to} $n_{iter}$ \textbf{do}
    
    \STATE \hspace{0.25cm} Select next sample $\mathbf{x}_n$ by \ethan{max}imizing$^*$ acquisition function $a(\mathbf{x})$:  \phantom{PPPPPPPPPPPPPPPPPPPPPPP} \phantom{PPP} $^*$\small{Maximization using Monte Carlo, See Section \ref{sec:monte} and Appendix \ref{app:monte}}
    \begin{align*}
        \mathbf{x}_{n}=\underset{\mathbf{x} \in \mathcal{X}}{\arg \max} \hspace{0.2cm} (a (\mathbf{x} ; y ) , \mathcal{D}_{n-1} )
    \end{align*}
    
    \STATE \hspace{0.25cm} Evaluate objective function at $\boldsymbol{x}_n$ and record $y_n$
    
    \STATE \hspace{0.25cm} Augment dataset: $ \mathcal{D}_{n} = \mathcal{D}_{n-1} \cup {\mathbf{x}_n, y_n }$
    
    \STATE \hspace{0.25cm} Retrain GP/DNO.
    \STATE \textbf{end for}
    \STATE \textbf{return} Final GP/DNO Model
    
  \end{algorithmic}
\end{algorithm}

\subsection{Surrogate Models}

We test two surrogate models in the BED framework, GPs as the benchmark case, and DNOs (specifically DeepONet) for greater scaling and generalization performance. While both GPs and DNOs provide the same role in the framework, their implementation is fundamentally different (see figure \ref{fig:ExpDesignSchematicDisc}~$b)$ for a visual representation). Whereas GPs are used to map random parameter inputs, \ethan{$\mathbf{x} \in \mathbb{R}^{1 \times D}$, where $D$ is the parameterization dimension, to a QoI $\tilde{G}(\mathbf{x})$, we use the DNOs to map a function to the QoI. 
In the case of the MMT problem, the input function varies spatially over $x$ as $u(x)=\mathbf{x} \boldsymbol{\Phi}(x)$, i.e. the scalar product between $\mathbf{x}$ and $\boldsymbol{\Phi}(x)$ where $\boldsymbol{\Phi}(x) \in \mathbb{R}^{D \times n_{x}}$ is the parameterization basis over $n_{x}$ spatial points,} to an output function $G(u(x)=\mathbf{x}\boldsymbol{\Phi}(x))$. In both cases, GP and DNO, $\mathbf{x}$ is the only independent variable and is maintained to provide a means of searching for extremes in the parameter space. However, the proposed regression task for DNOs ensures that the they perform their mapping in functional, or physical, space, rather than the parameter space. This distinction is foundational for our approach and success with DNOs for BED. 

We detail both GPs and DeepONet in detail in Appendices \ref{app:GP} and \ref{app:DON}, respectively, and only discuss our approach for quantifying the predictive variance, $\sigma^2(\mathbf{x})$, for DNOs using ensembles. 

\subsubsection{Ensemble of Neural Networks for Uncertainty Quantification} \label{sec:DNO_UQ}

Although neural network architectures are attractive for approximating nonlinear regression tasks, their complexity rids them of analytical expressions. This does not allow for a traditional Bayesian treatment of uncertainty in the underlying surrogate model -- a key property present for GP regression (see equation \ref{eqn:cov_GP}). Knowledge of the uncertainty of a surrogate model allows one to target model deficiencies as seen in the parameter space. This means that choosing an appropriate method for quantifying the uncertainty is a crucial and key component to active learning or BED. There are several techniques for quantifying uncertainty in neural networks and we provide a brief description of these in Appendix \ref{app:DON_Ensemble} and focus only on ensemble methods.

Ensemble approaches have been used extensively throughout the literature \cite{hansen1990neural,lakshminarayanan2016simple} and despite their improved results for identifying the underlying tasks at hand \cite{gustafsson2020evaluating}, their utility for quantifying uncertainty in a model remains a topic of debate. There are several approaches for creating ensembles. These include random weight initialization\cite{lakshminarayanan2016simple}, different network architectures (including activation functions), data shuffling, data augmentation, bagging, bootstrapping, and snapshot ensembles \cite{loshchilov2016sgdr,huang2017snapshot,smith2015no} among others. Here we employ random weight initialization, a technique found to perform similarly or better than Bayesian neural network approaches (Monte Carlo Dropout and Probablistic Backpropagation) for evaluation accuracy and out-of-distribution detection for both classification and regression tasks \cite{lakshminarayanan2016simple}. As stated earlier, much of the field is skeptical of the generality of ensembles to provide rigorous uncertainty estimates. However, recent studies, such as \cite{pickering2022structure} and specifically \cite{wilson2020bayesian}, have argued that DNN and DNO ensembles provide reasonable, if not superior, approximations of the posterior. Finally, the straightforward implementation of the randomly initialized weights motivates our choice, as it makes the adoption of these techniques far more probable.

We train $N$ randomly weight-initialized DNO models, each denoted as $\tilde{G}_{n}$, that find the associated solution field $y$ for functional inputs $u$ and operator parameters $z$ (i.e. components that change the underlying operator such as exponents or operator coefficients). This allows us to then determine the point-wise variance of the models as
\begin{equation}
    \sigma^2(u,z) = \frac{1}{(N-1)} \sum_{n=1}^{N} (\tilde{G}_{n}(u)(z) - \overline{\tilde{G}(u)(z)})^2,
\end{equation}
where $\overline{\tilde{G}(u)(z)}$ is the mean solution of the model ensemble. Finally, we must adjust the above representation to match the description for BED. In the case of traditional BED and GPs, the input parameters, $\mathbf{x}$, represent the union of two set of parameters, $\mathbf{x}_u$ and $\mathbf{x}_z$. The parameters $\mathbf{x}_u$ typically represent random variables applied to a set of functions that represent a decomposition of a random function $u = \mathbf{x}_u \mathbf{\Phi}(x_1, ... x_m)$ where $x_1, ..., x_m$ are discrete function locations (spatial, temporal, or both), while $\mathbf{x}_z = z$ represent non-functional parameters of the operator. Thus, the DNO description for uncertainty quantification may be recast as:
\begin{align}
    \sigma^2(\mathbf{x}) = \sigma(\mathbf{x}_u \cup \mathbf{x}_z) = \frac{1}{(N-1)} \sum_{n=1}^{N} (\tilde{G}_{n}(\mathbf{x}_u \mathbf{\Phi}(x_1, ... x_m))(\mathbf{x}_z) - \overline{\tilde{G}(\mathbf{x}_u \mathbf{\Phi}(x_1, ... x_m))(\mathbf{x}_z)})^2.
\end{align}


\subsection{Data Selection: Acquisition Functions} \label{sec:Acq}

The acquisition function is the key component of the sequential search algorithm, as it guides algorithm \ref{algorithm1} in exploring the input/parameter space and determines samples at which the objective function is to be queried. Because of the lack of a closed analytical form of DNOs, we only consider two acquisition functions used previously with GPs on several test cases in \cite{blanchard2021output}. The two functions we are interested in are the commonly used uncertainty sampling and the output-weighted uncertainty sampling proposed by \cite{blanchard2021output} and shown in \cite{sapsis22} to guarantee optimal convergence in the context of Gaussian Process Regression. The novelty here is that we apply them explicitly to DNOs (DeepONet) and present the advantages of DNOs compared to GPs as we apply them to a complex and high-dimensional problem.

\subsubsection{Uncertainty Sampling} \label{sec:US}

Uncertainty sampling (US) is one of the most broadly used active sampling techniques and identifies the sample where the predictive variance is the greatest, 
\begin{equation}
    a_{US}(\mathbf{x}) = \sigma^2(\mathbf{x}).
\end{equation}
Uncertainty sampling, also known as the active-learning-MacKay (ALC) algorithm \cite{gramacy2009adaptive}, imposes a sequential search that evenly distributes uncertainty over the input space as it gains data. The popularity of US is due to three qualities: ease of implementation, inexpensive evaluation (for small datasets with GPs), and analytic gradients, the last of which permits the use of gradient-based optimizers.

\subsubsection{Likelihood-Weighted Acquisition Functions} \label{sec:LW}

There are several ``extreme event'' likelihood-weighted acquisition functions that could be explored, as proposed by \cite{blanchard2021bayesian}, but we elect to only test the likelihood-weighted uncertainty sampling (US-LW) acquisition because of its simplicity in implementation. For US-LW, we augment the US sampling acquisition function with the previously described danger scores to give 
\begin{equation}
    a_{US-LW}(\mathbf{x}) = w(\mathbf{x}) \sigma^2(\mathbf{x}) ,
\end{equation}
such that both \textit{highly uncertain} and \textit{high magnitude} regions are sampled.


To compute $ w(\mathbf{x})$, we note that the approximated output PDF, $p_\mu(\mu)$ is approximated via a kernel density estimator with $n=10^6$ test points ($10^7$ for the $20D$ example). For the DNO cases, we chose to compute this with only the first ensemble member, $\mu = \tilde{G}_{1}$, to reduce computational costs. Similar to the ensemble results for $N=2$, using only one ensemble member is akin to using Thompson sampling \cite{thompson1933likelihood,chapelle2011empirical} and performs without reduction in performance.

\subsection{QoI and Log-PDF Error Metrics} \label{sec:danger_metrics}

To test the ability of the DNO and GP Bayesian-inspired sequential algorithms to quantify extremes, we define a quantity of interest, (QoI), or ``Danger Map'', for the pandemic scenarios and rogue waves. The rogue wave QoI is defined as,
\begin{equation}
    G(\mathbf{x}) = || Re(u(x,t=T;\mathbf{x}))||_{\infty},
\end{equation}
where $T=20$, while the pandemic QoI is,
\begin{equation}
    G(\mathbf{x}) = I(t=T;\mathbf{x}),
\end{equation}
where $T=45$ days.

\ethan{For each case, we then select $10^5$ LHS test samples, $\mathbf{X} \in \mathbb{R}^{d \times 10^5}$, evaluate the true QoI at each, $\mathbf{y} = G(\mathbf{X}) \in \mathbb{R}^{1 \times 10^5}$, compute the probability of each sample, $\mathbf{\alpha} = p_\mathbf{x}(\mathbf{X})\in \mathbb{R}^{1 \times 10^5}$ and find the true PDF, $p_{G}(y) = \mathrm{KDE}(\mathrm{data}=\mathbf{y}, \mathrm{weights}=\mathbf{\alpha})$, using standard gaussian KDE implementations (e.g. \textsc{scipy.stats.{gaussian\_kde}}). The approximated PDF is then found by replacing the true map with the surrogate map at each iteration.}


\ethan{While the $10^5$ test points provide sufficient samples for accurate PDF assessment through a KDE at all cases $10D$ and less, the absolute truth PDF for the $20D$ example is much more difficult. To attain a definitively converged PDF would require $>10^7$ samples, a computationally infeasible quantity. Therefore, we emphasize that our truth metric is based upon how well the approximated $10^5$ test points reconstruct the KDE PDF given the same true $10^5$ test points, rather than a definitive converged truth. This means that as long as the true behavior and the surrogate model are the same on these $10^5$ Latin hypercube samples, then the PDFs generated are identical everywhere. Although this may seem like a simple task, our results only sample a maximum of 5000 training points (see figure \ref{fig:Lines_LHS}), two orders of magnitude less than the test set. At this size, only the DNOs with output weighted sampling are unable to accurately regress to the test set. Therefore, this shows that the method is able to learn the underlying map at a substantially improved efficiency.}

Finally, to determine whether the testing data appropriately identifies extremes, we compute the log-PDF error
\begin{equation}
\label{eq:log-pdf-error}
 e(n) = \int | \log_{10} p_{\mu_n} (y) - \log_{10} p_G (y)| \text{d} y. 
\end{equation}



\ethan{For the LAMP problem, where the output measure (VBM) is a time series, the QoI and the error metric are slightly more complicated. The 
GP and DNO map $\mathbf{x}$ to $ \mathbf{q}$ as,
\begin{equation}
    G(\mathbf{x}) = \mathbf{q} (\mathbf{x}),
\end{equation}
and the VBM in time is recovered via 
\begin{equation}
    y(t;\mathbf{x}) = \sum_{i=1}^{12} q_i(\mathbf{x}) \phi_{q_i}(t),
\end{equation}
where $\phi_{q_i}(t)$ are output basis vectors in time. We then concatenate realizations of $y(t; \mathbf{x})$ to form a single, long time series $Y(t)$.  We drop explicit dependence of $Y(t)$ on the $\mathbf{x}$, under ergodicity assumptions.  Finally, our quantity of interest is the one-point time statistics of $Y(t)$. The ground truth pdf $p_y(y)$, is computed using 3000 Monte Carlo realizations each over 1800 time units, while the surrogate model approximation, $p_{\mu_n}(y)$ uses 10000 LHS realization of 40 time units. Both PDFs are computed using standard unweighted KDE and the log-PDF error is computed just as in equation \eqref{eq:log-pdf-error}.} 












\subsection{Monte Carlo Optimization of Acquisition Functions} \label{sec:monte}

In our experiments, we consistently observe that acquisition samples found through optimizers using gradient descent are not globally optimal. Instead, Monte Carlo evaluation of the DNOs and GPs consistently find superior optima. This is chiefly because of the non-convexity of the acquisition function. We may recall the highly non-convex behavior of the 2D acquisition fields in figure~\ref{fig:PandemicExample}~$b)$, even with as little as $\approx 5$ samples. This non-convex nature emits many local minima that require many initial search samples to provide confidence that the chosen optima are nearly global. As the optimizer progresses for each iteration, it must call the DNO or GP, whereas a Monte Carlo approach may efficiently evaluate all samples in one vector operation. This means that for the same computation time as the optimizers, Monte Carlo sampling may evaluate a substantially larger distribution of query samples and return acquisition samples with superior scores than that of the optimizer. In Appendix \ref{app:monte} we show that acquisition scores found via Monte Carlo at $20D$ consistently outperform optimizers for similar computation times.

\ethan{While we choose to implement a Monte Carlo approach instead of off-the-shelf optimizers, there are likely several other optimization approaches that would prove to be superior in finding optimal sets of acquisition points. However, our contribution is focused on defining and implementing the acquisition function in DNOs for \textit{selecting the next experiment} and leave further optimization of the DNO acquisition space for future work.}

\subsection{Experiment Batching} \label{sec:batch}

As systems become more complex, additional experiments/data are required to reduce errors for higher dimensional cases, as observed in figure~\ref{fig:1d_3D_Errors}. Considering many experiments can be conducted in parallel, we ask whether choosing multiple local minima of the acquisition function presents marginally reduced performance than a purely sequential search. This is especially critical for situations where experimental time is more costly than additional setups (e.g. protein or genetic design).

The purpose of batching is to find \textit{multiple regions of local optima} of the acquisition function, rather than finding \textit{several optima in the same region}. To impose this idea, we create a constraint that no acquisition sample may reside closer than a distance $r_{\min}$ to each other. We define  $r_{\min}$ as a fraction of the maximum euclidean distance of the space being sampled,
\begin{equation}
    r_{\min} =  r_{l} \bigg( \sum_{d=1}^{D} (x_{d,+} - x_{d,-})^2\bigg)^{2},
\end{equation}
where $x_{d,+}$ and $x_{d,-}$ are the maximum and minimum domain bounds of each parameter dimension $d$ and $r_{l}$ is the user defined percentage. In this work, we chose a static $r_{l} = 0.025$, but dynamic values based on the packing of the parameter space would be an intriguing direction for increasing the efficacy of this approach. Imposing this constraint requires an iterative processing of the acquisition scores, detailed in Algorithm \ref{algorithm2}. For the batching applied in this study, each case uses a Monte Carlo querying of $n_{q} = 10^6$ points. 
\begin{algorithm}
  \caption{Sequential selection of samples for batch sampling.}\label{algorithm2}
  \begin{algorithmic}[1]
    \STATE \textbf{Evaluate} acquisition function for $n_{q}$ query points. 
    \STATE \textbf{for} Acquisition samples smaller than batch size $n_{a} < n_{b}$
    \STATE \hspace{0.25cm} \textbf{Choose} the maximum score, $\max(a) = a(\mathbf{x}_{c})$, from $n_{q}$ points.
    \STATE \hspace{0.25cm} \textbf{Augment} $\mathbf{x}_{a}$ with chosen point, $\mathbf{x}_{c}$. 
    \STATE \hspace{0.25cm} \textbf{Compute} the distances, $r$, between the chosen point and the remaining query points.
    \STATE \hspace{0.25cm} \textbf{Eliminate} all samples from $n_{q}$ where $r$ < $r_{\min}$.
    \STATE \textbf{end for}
    \STATE \textbf{return} samples for the next experiment: $\mathbf{x}_{a}$
  \end{algorithmic}
\end{algorithm}



\subsubsection*{Code and Data Availability}
If accepted, all data and code will be shared via an open source link.


\subsubsection*{Acknowledgments} The authors acknowledge support from DARPA (Grants HR00112290029 and HR00112110002), as well as the AFOSR MURI Grant FA9550-21-1-0058 and the ONR Grants N00014-20-1-2366 and N00014-21-1-2357,  awarded to MIT. They also grateful to Dr. Vadim Belenky and Mr. Ken Weems from NSWC at Carderock for support on the LAMP code.

\bibliographystyle{abbrv}
\bibliography{references, sj_bib}


\appendix

\section{SIR Pandemic Model} \label{app:SIR}

We implement a simple Susceptible, Infected, Recovered (SIR) model proposed by \cite{kermack1927contribution} and reintroduced by \cite{anderson1979population},
\begin{align}
    \frac{\mathrm{d}S}{\mathrm{d}t} &= -\beta IS + \delta R \\
    \frac{\mathrm{d}I}{\mathrm{d}t} &= \beta IS - \gamma I \\
    \frac{\mathrm{d}R}{\mathrm{d}t} &= \gamma I - \delta R,
\end{align}
where $\delta$ is the rate of immunity loss, $\gamma$ is the recovery rate, and $\beta$ is the infection rate. Here we take $\delta = 0$ and $\gamma=0.1$ and adjust $\beta$ from a scalar to a stochastic infection rate, $\beta(t)$, defined as
\begin{equation}
    \beta(t) = \beta_0 (\mathbf{x} \mathbf{\Phi}(t) + \phi_0),
\end{equation}
where $\Phi(t)$ is found via a Karhunen-Loeve expansion of a radial basis kernel with $\sigma^2_{\beta} = 0.1$ and length scale $\ell_{\beta} = 1$, $\beta_0 = 3 \times 10^{-9}$, and $\phi_0 = 2.55$, to ensure all infection rates are non-negative. Initial conditions for the model are $I_0 = 50$ with total population $P = 10^8$, and a step size of 0.1 days is used over 45 days.

\section{Dispersive Nonlinear Wave Equation: Majda, McLaughlin, and Tabak Model} \label{app:NLS}

\begin{figure}
\centering
$a)$ \hspace{6cm} $b)$ \phantom{PPPPPPPPPPPPPPPPPPPPPPPPPPPPP}
\includegraphics[width=1\textwidth,trim={0cm 0cm 0cm 0cm},clip]{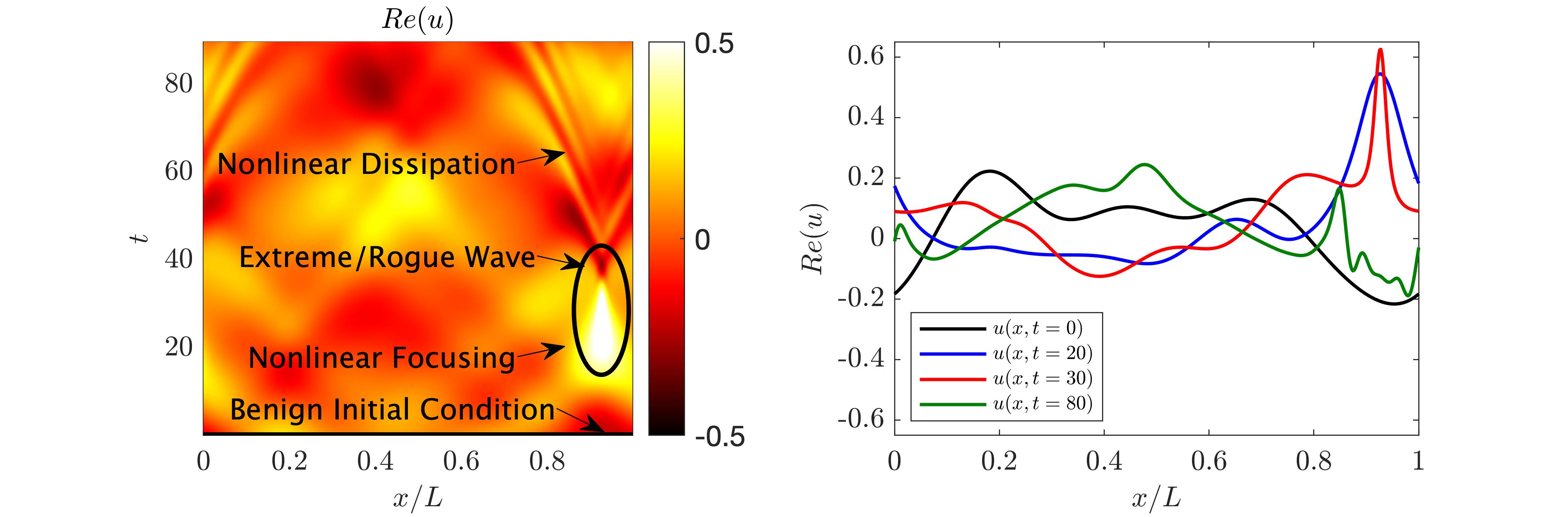}
\caption{\textbf{The MMT emits many nonlinear phenomena in finite time, from focusing to dispersion.}$a)$ A realization of the MMT model ($\alpha=1/2,\beta=0,\lambda=-0.5$) we discover with DNO-BED, where a benign initial condition leads to an extreme, rogue wave. $b)$ Wave height of selected times from $a)$.}
\label{fig:rogue_wave}
\end{figure}

The Majda, McLaughlin, and Tabak \cite{majda1997one} (MMT) model is a dispersive nonlinear wave equation used for studying $1 \mathrm{D}$ wave turbulence.  Under appropriate choice of parameters it is associated with the formation of intermittent events \cite{cousins2014quantification}. It is described by
\begin{equation}
    i u_{t}=\left|\partial_{x}\right|^{\alpha} u+\lambda\left|\partial_{x}\right|^{-\beta / 4}\left(\left.\left.|| \partial_{x}\right|^{-\beta / 4} u\right|^{2}\left|\partial_{x}\right|^{-\beta / 4} u\right)+i D u,
    \label{eqn:MMT}
\end{equation}
where $u$ is a complex scalar, exponents $\alpha$ and $\beta$ are chosen model parameters, and $D$ is a selective Laplacian (described further below). This model gives rise to four-wave resonant interactions that, especially when coupled with large scale forcing and small scale damping, produces a family of spectra revealing both direct and inverse cascades \cite{majda1997one,cai1999spectral}. A realization of the MMT model is shown in figure~\ref{fig:rogue_wave} that demonstrates these complex dynamical properties. Not only does this model provide a rich dynamical response, but also presents a unique utility as a physical model for extreme ocean waves, or rogue waves \cite{zakharov2001wave,zakharov2004one,pushkarev2013quasibreathers}. Hence, its study is an ideal test bed for both examining the numerical difficulties of predictive models and uncovering insights to physical, real-world applications.


For both ease of computation and for discussion of the terms in the MMT model, we transform the equation into wavenumber space. The pseudodifferential operator $\left|\partial_{x}\right|^{\alpha}$, via the Fourier transform in space becomes: $\widehat{\left|\partial_{x}\right|^{\alpha} u(k)}=|k|^{\alpha} \widehat{u}(k)$ where $k$ is the wavenumber in $x$. This formulation may be similarly defined on a periodic domain. We choose $\alpha = 1/2$ and $\beta = 0 $ as done in \cite{cousins2014quantification}, reducing equation \eqref{eqn:MMT} to
\begin{equation}
     \widehat{u}(k)_{t}=-i|k|^{1/2} \widehat{u}(k)-i \lambda| \widehat{u}(k)|^{2}  \widehat{u}(k) +\widehat{D u}(k) + f(k),
    \label{eqn:MMT_a05_b0}
\end{equation}
where $f(k)$ is a forcing and $\widehat{D u}(k)$ is a selective Laplacian of the form:
\begin{equation}
    \widehat{D u}(k)=\left\{\begin{array}{ll}-\left(|k|-k^{*}\right)^{2} \hat{u}(k) & |k|>k^{*} \\ 0 & |k| \leq k^{*}\end{array}\right.
    \label{eqn:selective_Lap}
\end{equation}
where $k^{*}$ presents the lower bound of wavenumbers subject to dissipation. For the model considered in this study we choose $\lambda = -0.5$, $k^* = 20$, $f(k) = 0$, $dt = 0.001$, and a grid that is periodic between 0-1 with $N_x = 512$ grid points. To propose a stochastic and complex initial condition, $u(x,t=0)$, we take the complex-valued kernel
\begin{equation}
 k(x,x^\prime)   = \sigma_{u}^2 e^{i(x-x^\prime)} e^{-\frac{(x-x^\prime)^2}{\ell_u}},
\end{equation}
with $\sigma_u^2 = 1$ and $\ell_u = 0.35$. We then parametrize the stochastic initial conditions by a finite number of random variables, $\mathbf{x}$, using the Karhunen-Loeve expansion of the kernel's correlation matrix,
\begin{equation}
    u(x,t=0) \approx \mathbf{x} \boldsymbol{\Phi}(x), \hspace{0.5cm} \forall \hspace{0.5cm} x \in [0,1)
\end{equation}
where $\mathbf{x} \in \mathbb{C}^m$ is a vector of complex coefficients and both the real and imaginary components of each coefficient are normally distributed with zero mean and diagonal covariance matrix $\Lambda$, and $ \{ \boldsymbol{\Lambda}, \boldsymbol{\Phi} (x)\} $ contains the first $m$ eigenpairs of the correlation matrix. This gives the dimension of the parameter space as $2m$ to \ethan{account for the real and imaginary components of each coefficient}. The presented $2D, 4D, 6D, 20D$ results correspond to $m=1,2,3,10$. For all cases, the random variable $x_i$ is restricted to a domain ranging from -6 to 6, in that 6 standard deviations in each direction from the mean. 

\ethan{\section{Efficient Estimation of Structural Fatigue for Ship Design} \label{app:Lamp}}

\ethan{The LAMP-based problem aims to efficiently estimate the VBM statistics of a unique ship, the ONR topsides flare variant, to inform future design decisions. This real-life industrial application brings significant complexity and numerous challenges. These are documented at length in Guth and Sapsis (2022) \cite{guth22} and we strongly refer the reader to the above paper as we closely follow their implementation. Briefly, we describe the inputs and outputs below.}

\ethan{Similar to the previous problems, we parametrize a low-dimensional projected subspace of wave episodes using a Karhunen Lo\`{e}ve series expansion \cite{karhunen47,loeve48} of the stochastic surface waves, described by a power spectral density such as the JONSWAP spectrum.  
We next use the proprietary code LAMP (Large Amplitude Motion Program, v4.0.9, May 2019) \cite{lin10} in order to calculate the forward problem of specific VBM response to a specific wave episode.  A specific wave episode, described by the parametrization $\mathbf{x}$, is mapped onto a VBM time series $y(t)$, whose duration is related to the original interval of the Karhunen Lo\`{e}ve expansion.}

\ethan{Finally, as the LAMP code is proprietary and cannot be shared, we perform our active search in a precomputed dataset of 3000 (expensive simulations) LHS samples in $10D$, with each input variable varying from $[-4.5,4.5]$. This data is provided with the code.}

\section{Gaussian Process Regression} \label{app:GP}

For low-dimensional problems, Gaussian process (GP) regression \cite{rasmussen2003gaussian} is the ``gold standard’’ for Bayesian design. There are several attractive qualities of GPs. They are agnostic to the details of a black-box process (just like neural networks described next) and they clearly quantify both the uncertainty in the model and the uncertainty associated with noise. GPs are also relatively easy to implement and cheap to train.

A Gaussian process $\bar{f}(\mathbf{x})$, where $\mathbf{x}$ is a random variable, is completely specified by its mean function $m(\mathbf{\mathbf{x}})$ and covariance function $k\left(\mathbf{\mathbf{x}}, \mathbf{\mathbf{x}}^{\prime}\right)$. For a dataset $\mathcal{D}$ of input-output pairs ($\{\mathbf{\mathbf{x}}, \mathbf{y}\})$ and a Gaussian process with constant mean $m_{\boldsymbol{0}}$, the random process $\bar{f}(\mathbf{\mathbf{x}})$ conditioned on $\mathcal{D}$ follows a normal distribution with posterior mean and variance
\begin{equation}
\mu(\mathbf{\mathbf{x}})=m_{0}+k(\mathbf{\mathbf{x}}, \mathbf{\mathbf{x}}) \mathbf{K}^{-1}\left(\mathbf{y}-m_{0}\right)
\label{eqn:mean_GP}
\end{equation}
\begin{equation}
\sigma^{2}(\mathbf{\mathbf{x}})=k(\mathbf{\mathbf{x}}, \mathbf{\mathbf{x}})-k(\mathbf{\mathbf{x}}, \mathbf{\mathbf{x}}) \mathbf{K}^{-1} k(\mathbf{\mathbf{x}}, \mathbf{\mathbf{x}})
\label{eqn:cov_GP}
\end{equation}
respectively, where $\mathbf{K}= k (\mathbf{\mathbf{x}}, \mathbf{\mathbf{x}})+\sigma_{\epsilon}^{2} \mathbf{I}$. Equation \eqref{eqn:mean_GP} can predict the value of the surrogate model at any sample $\mathbf{\mathbf{x}}$, and \eqref{eqn:cov_GP} to quantify uncertainty in prediction at that sample \cite{rasmussen2003gaussian}. Here, we chose the radial-basis-function (RBF) kernel with automatic relevance determination (ARD),
\begin{equation}
k\left(\mathbf{\mathbf{x}}, \mathbf{\mathbf{x}}^{\prime}\right)=\sigma_{f}^{2} \exp \left[-\left(\mathbf{\mathbf{x}}-\mathbf{\mathbf{x}}^{\prime}\right)^{\top} \mathbf{L}^{-1}\left(\mathbf{\mathbf{x}}-\mathbf{\mathbf{x}}^{\prime}\right) / 2\right],
\end{equation}
where $\boldsymbol{L}$ is a diagonal matrix containing the lengthscales for each dimension and the GP hyperparameters appearing in the covariance function $(\sigma_{f}^{2}$ and $\boldsymbol{L}$ in \eqref{eqn:cov_GP} are trained by maximum likelihood estimation).

One setback from the above expression is the inference step in GP regression, where each iteration requires the inversion of the matrix $\mathbf{K}$. Typically performed by Cholesky decomposition, the inversion cost scales as $O(n^{3})$, with $n$ being the number of observations \cite{rasmussen2003gaussian,shahriari2015taking}. This means that as problems grow to infinite dimensions, inevitably requiring larger datasets, GPs become prohibitively costly. We will show here that as the need for data increases, GPs become much more computationally intensive than our next surrogate model, neural networks.

\section{Deep Neural Operators and DeepONet} \label{app:DON}

Unlike GPs, Deep Neural Operators or Deep Neural Networks, do not suffer from data-scaling challenges and are our primary model class for consideration in this manuscript. Deep Neural Networks, when cast as neural operators \cite{kovachki2021neural}, are specifically well-suited for characterizing infinite dimensional systems, as they may map functional inputs to functional outputs. Although our work is general for any neural network approach, we leverage the architecture proposed by \cite{lu2021learning} for approximating nonlinear operators: \textit{DeepONet}.


DeepONet seeks approximations of nonlinear operators by constructing two deep neural networks, one representing the input function at a fixed number of sensors and another for encoding the ``locations'' of evaluation of the output function. The first neural network, termed the ``branch'', takes input functions, $u$, observed at discrete sensors, $x_i, i = 1 ... m$. These input functions can take on several representations, such as initial conditions (i.e. $u_0$) or forcing functions. The second neural network, termed the ``trunk'', should be seen as an encoder for inherent qualities of the operator, denoted as $z$ (referred to as $y$ in \cite{lu2021learning}, but changed here for typical active sampling notation). For example, a variable coefficient or exponent in the true nonlinear operator, $G$, alternatively, and the usual use case for DeepONet, the trunk variable can refer to the evaluation of the operation to an arbitrary point in time and/or space. Together, these networks seek to approximate the nonlinear operation upon $u$ and $z$ as $G(u)(z) = y $, where $y$ will denote the scalar output from the $u,z$ input pair. 

Given a set of input-output pairs, $\{[\mathbf{u},\mathbf{z}], G(\mathbf{u})(\mathbf{z})\}$, DeepONet seeks to minimize the difference between the true operator, $ G(u)(z)$, and the dot product between two neural networks $\mathbf{g}(u)$ and $\mathbf{f}(z)$. These network's level of expressivity is governed by the number of neurons ($n$), the number of layers ($l_b, l_t$ for branch and trunk, respectively), and activation functions. Under sufficient training DeepONet can meet any arbitrary error, $\epsilon$, as  
\begin{equation}
\mid G(u)(z)-\langle\underbrace{\mathbf{g}\left(u\left(x_{1}\right), u\left(x_{2}\right), \cdots, u\left(x_{m}\right)\right)}_{\text {branch }}, \underbrace{\mathbf{f}(z)}_{\text {trunk }} \rangle|<\epsilon.
\end{equation}
Thus, DeepONet can, to an arbitrarily small precision, approximate an infinite dimensional nonlinear operator $G(u)(z)$. However, the functions $\mathbf{g}$ and $\mathbf{f}$ are typically trained under the assumption of plentiful data. We are interested in how DeepONet can be optimally trained with the least amount of data to discover and quantify extreme events.

DeepONet is attractive for these tasks as neural nets generalize well and are incredibly fast to evaluate (compared to their experimental/simulation counterparts) for arbitrarily chosen points. However, there is little work on how one optimally selects the best samples to train a neural net. Previously, the approach taken has included creating an appropriate basis for specific operators that are used to train DeepONet \cite{lu2019deeponet}. For the problems we are interested in (i.e. rare and extreme events), this appropriate basis is unknown. Therefore, we envision DeepONet will provide a flexible model to learn seen data, and then provide early predictions of where danger and uncertainty lie in the input/output space through DeepONet’s parameterization, that we may then query the underlying system to best inform DeepONet.

\subsection{Approaches for quantifying uncertainty in Deep Neural Networks}  \label{app:DON_Ensemble}

There are several techniques for quantifying uncertainty in neural networks and three categorizations of these techniques we wish to mention: single deterministic networks, Bayesian neural network inference approaches, and ensemble methods (see \cite{gawlikowski2021survey} and \cite{psaros2022uncertainty} for comprehensive reviews of these methods for uncertainty quantification). Unfortunately, a complex array of advantages and disadvantages of each approach provides no clear favorite. \ethan{Single model, i.e. deterministic, methods} \cite{sensoy2018evidential,malinin2018predictive} use one forward pass of a deterministic network to learn both the mean and variance of a labeled output. Using only one model leads to cheap training and evaluation. \ethan{Unfortunately,} this single opinion of the underlying system results in substantial sensitivity, an unattractive quality for regression problems quantifying physical instabilities/extreme events. Bayesian neural networks \cite{blundell2015weight,gal2016dropout,gal2017deep,wilson2020bayesian} encompass a broad variety of stochastic DNNs that combine Bayesian inference theory with the expressiveness, scalability, and predictive performance enjoyed by deep neural networks. Although the supporting theory behind Bayesian methods, as well as empirical results, imparts faith that such models will lead to the greatest chance of success, they do come with a significant disadvantage, complexity.  Bayesian neural networks are significantly more complex than standard neural networks and can be exhausting to train, making them difficult to implement. From an academic viewpoint, this can be overcome. However, as a study concerned with translating DNNs and active learning to practical engineering systems that solve real-world problems, we opt for using the straight-forward ensemble approach. An approach that seats itself between the single deterministic model and the infinite model representations of Bayesian neural networks. 

\section{DeepONet Setup} \label{app:hyper}

\begin{figure}
\centering
\includegraphics[width=0.32\textwidth,trim={0cm 0cm 0cm 0cm},clip]{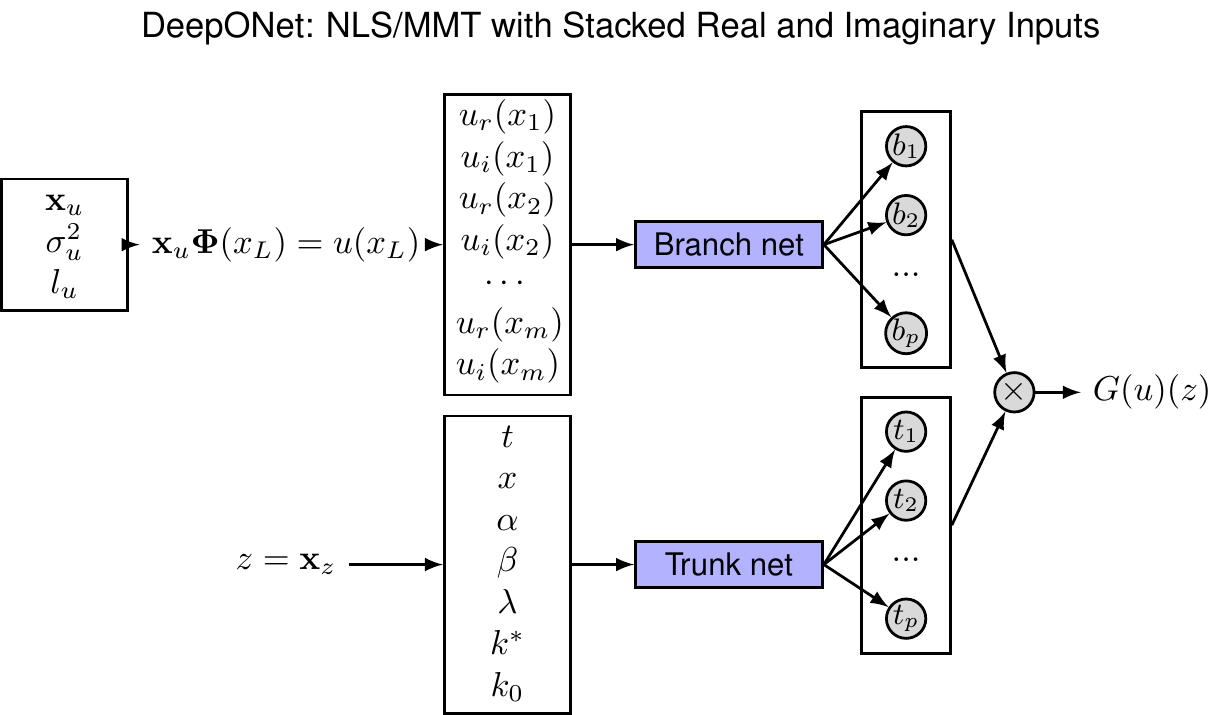}
\includegraphics[width=0.32\textwidth,trim={0cm 0cm 0cm 0cm},clip]{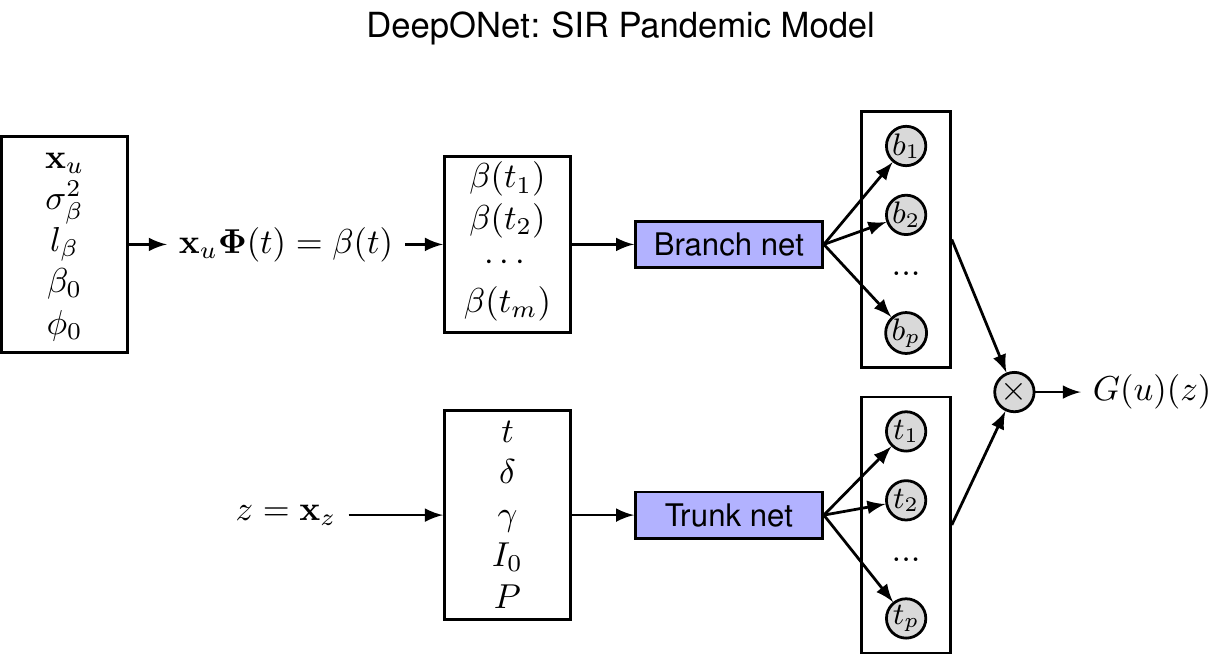}
\includegraphics[width=0.32\textwidth,trim={0cm 0cm 0cm 0cm},clip]{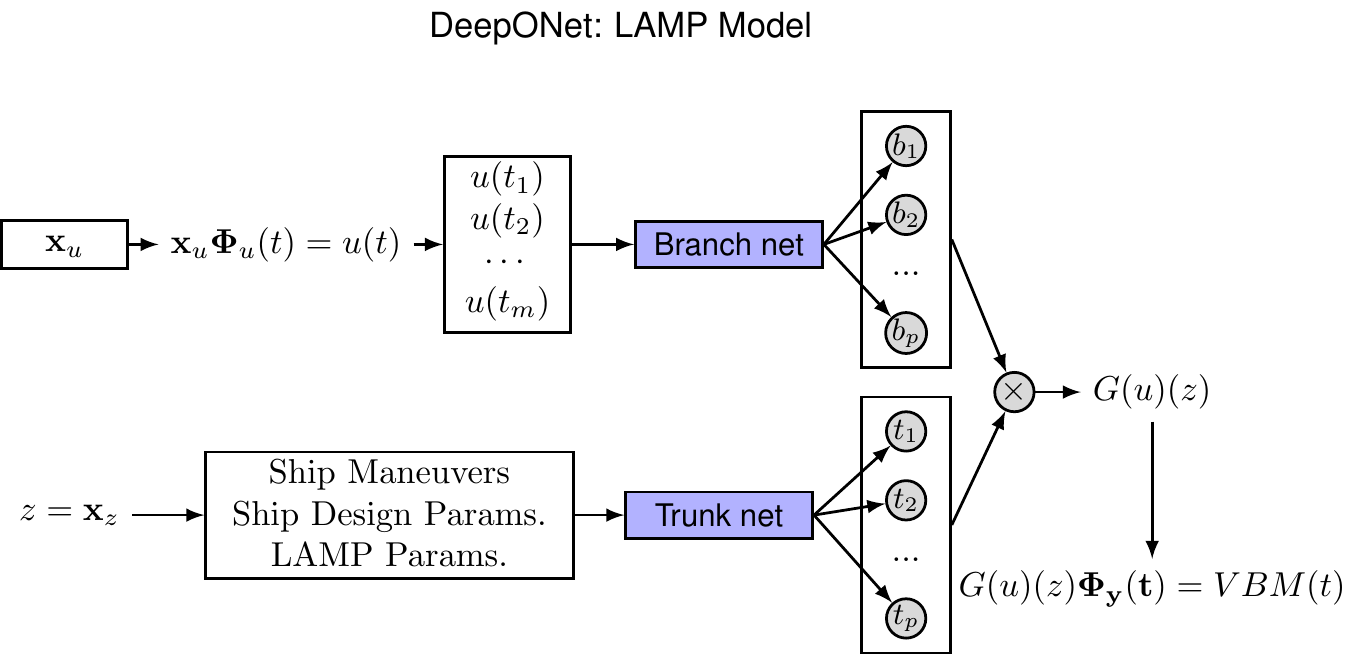}
\caption{\textbf{The application of the MMT, SIR, and LAMP operators to DeepONet.}}
\label{fig:DON}
\end{figure}

Application of the MMT, SIR, and LAMP models to the DeepONet architecture requires a well-posed distinction between the functions and parameters that belong in the branch and trunk networks. Figure \ref{fig:DON} provides a visual of the function/parameter delineations and the architecture for each modeling task. 


In this work, we provide all input initial conditions as functions to the branch network. Because of the requirement that all inputs to DeepONet take on real values, the MMT setup requires that we split the input function, $u$, into its real ($u_r$) and imaginary ($u_i$) components. It is critical to keep both components, as each contains unique and coupled information that is propagated in the MMT operation (this is not necessary for the real-valued SIR or LAMP model). We then stack these components as a vector by $x$ position/sensor with the imaginary component directly following the real component at each $x$ position. Technically, the ordering of the inputs does not matter, due to the linear nature of the first layer, but the ordering of the inputs must remain consistent. For increased speed, we also reduce the sensor number of $x_m$ sensors from the 512 used for the direct MMT calculation to 128 points for DeepONet (this corresponds to 256 total inputs values because of the real and imaginary components), 125 points for the SIR infection rate, and 1024 points for the LAMP wave episode. We also give the length scale and variance as input parameters, where applicable, as they define the kernel that emits the Karhunen-Loeve expansion functions. The input functions directly recognize adjustments to these parameters. 

The trunk contains parameters that are intrinsic to the MMT, SIR, and LAMP operations. For MMT, these comprise both time and space ($t,x$) and the chosen constants $\alpha,\beta,\lambda,k^*$. Regarding BED, we may assign each of these parameters to an appropriate prior distribution, such as uniform for space and time. In this study, we choose all values to remain constant and although the trunk is straightforward to implement, it is simply a redundant network here. 


We provide the hyperparameters and other quantities used for the various MMT cases performed throughout the manuscript in table~\ref{tab:hyper}. In addition to the hyperparameters, we use the ReLu activation function, a learning rate of $l_r = 0.001$, and 1000 epochs for each training procedure. For the SIR search, the same hyperparameters are used with 8 branch layers, 1 redundant trunk layer, and 300 neurons. For the LAMP search, the same hyperparameters are used with 5 branch layers, 1 redundant trunk layer, and 200 neurons.

\begin{table}
\centering
\begin{tabular}{c c c c c }  
 Case & Neurons & Branch Layers & Ensemble Size & Experiments \\  \hline
 MMT: $2-6D$ & 200 & 5 & 10 & 10 \\ 
 MMT: $8D$ Batching & 200 & 5 & 10 & 10 \\ 
 MMT: $8D$ Ensemble, $n \leq 2500$ & 200 & 5 & 2,4,8,16 & 10 \\ 
 MMT: $8D$ Ensemble, $n  > 2500$ & 200 & 6 & 2,4,8,16 & 10 \\ 
 MMT: $20D$, $n \leq 2500$  & 200 & 5 & 2 & 25 \\ 
 MMT: $20D$,  $n > 2500$  & 200 & 6 & 2 & 25 \\ 
 LAMP: $10D$  & 200 & 5 & 2 & 10 \\ 
 SIR: $2D$  & 300 & 8 & 2,8 & 1 \\ 
\end{tabular}
\caption{\textbf{Hyperparameters used for all cases.} All trunk layers $=1$ due to their redundancy.}
\label{tab:hyper}
\end{table}

\section{Monte Carlo Optimization at $20D$}\label{app:monte}

Here we show that acquisition samples found through Monte Carlo (MC) searches are consistently superior than those selected by optimizers using gradient descent algorithms (when reasonably similar computation times are considered). To demonstrate this behavior, we test four methods for finding 50 batched acquisition samples for the $20D$ MMT case at 500, 2500, and 5000 training samples, $N=2$ ensemble members, and over 25 independent experiments. The four approaches are listed below, with each case choosing the 50 best samples with algorithm \ref{algorithm2}:
\begin{enumerate}
    \item L-BFGS-B algorithm, implemented within the \texttt{scipy Python} package, $10^2$ random LHS initial points. 
    \item MC with $10^5$ random uniform initial samples (sampled from a uniform distribution for speed)
    \item MC with $10^5$, best 100 samples (algorithm \ref{algorithm2}) passed to L-BFGS-B.
    \item MC with $10^6$ random uniform initial points.
\end{enumerate}

Table \ref{tab:Times} and table \ref{tab:scores} provide the mean computation times and scores for all cases, respectively. The mean computation times ($\pm$ one standard deviation) are clearly faster for MC methods, by 5-fold, compared to any approach with the L-BFGS-B algorithms. We also note that only the first ensemble member is used for computing and querying $p_{\mu}$ for speed, as well as likely better performance. 

\begin{table}
\centering
\begin{tabular}{c||c|c|c|c}
    \textbf{Samples} &  \phantom{P}  L-BFGS-B \phantom{P} & \phantom{PP} MC: $10^5$  \phantom{PP} & MC+L-BFGS-B &  \phantom{PP} MC: $10^6$  \phantom{PP} \\
    \hline
    500 & $122 \pm 7 $ &  $3.6 \pm 0.1$ &	$123 \pm 5$ &	$\mathbf{27 \pm 1}$ \\
    2500 & $126 \pm 6$ &  $4.0 \pm 0.2$ &	$125 \pm 5$ &	$\mathbf{26 \pm 2}$ \\
    5000 & $127 \pm 8$ &  $3.7 \pm 0.1$ &	$135 \pm 7$ &	$\mathbf{27 \pm 3}$ \\
\end{tabular}
\caption{\textbf{MC is faster than built-in optimizers.} Mean compute times ($k=25$) for batch minimization ($n_b=50$) of the $20D$ US-LW acquisition function with $\pm$ one standard deviation. Best values are bold.}
\label{tab:Times}
\end{table}

Table \ref{tab:scores} provides the mean scores (over all chosen batch samples and experiments, $n=1250$), as well as the mean best and worst sample for each experiment ($n=25$). It is critical to note that at these high dimensions the difference between acquisition scores can be \textit{hundreds} of orders of magnitude different (this is a direct result of weights define by the ratio of two PDFs computed over 20 dimensions). To navigate this computational challenge, we take the $\log_{10}$ of the scores and apply a negative sign such that lower scores are optimal (i.e. $-\log_{10}(a(\mathbf{x})))$. Considering the mean of all chosen points, we see that a naive use of the L-BFGS-B optimizer finds scores that are $\approx 50$ orders of magnitude worse than the $10^6$ MC approach, regardless of the training complexity of the model. The L-BFGS-B optimizer only provides a marginal advantage for finding the best sample in any random instance, however, the standard deviation is 30 orders of magnitude and suggests the approach is terribly inconsistent. Juxtaposing the min and max results, we see that the MC method only varies by approximately 10 orders of magnitude compared to the 80 orders of the L-BFGS-B.

Overall, the MC method provides greater consistency in finding several attractive acquisition points, while also enjoying ease of implementation and better computational efficiency. Of course, if given enough computational resources and time, the L-BFGS-B method permits pinpointing exceptional acquisition points. Approaches combining both MC and L-BFGS-B, where the MC samples are further optimized (as also implemented here) or performed from more LHS points, can either refine or enrich the set of acquisition points.

We believe the results found here are chiefly because of the non-convexity of the acquisition function. We may recall the highly non-convex behavior of the 2D acquisition fields in figure~\ref{fig:PandemicExample}$b)$, even with as little as $\approx 10$ samples. This non-convex nature emits many local minima that require many initial samples to provide confidence that the chosen optima are nearly global. This makes the task for built in optimizers extremely difficult, especially for high dimensional problems, and lead to optimizations that become easily fooled or stuck. This difficulty is so challenging that the sparse MC sampling of only $10^6$ points, an extremely sparse sampling of a $20D$ space, easily outperforms the optimizers.

\begin{table}
\centering
\begin{tabular}{c||c|c|c|c}
    \textbf{Samples} &  \phantom{P}  L-BFGS-B \phantom{P} & \phantom{PP} MC: $10^5$  \phantom{PP} & MC+L-BFGS-B &  \phantom{PP} MC: $10^6$  \phantom{PP} \\
    \hline
    Mean ($k=1250$)& \\
    \hline
    500  & $110 \pm 18$ & $70 \pm 5$	        & $66 \pm 6$    & $\mathbf{60 \pm 4} $ \\
    2500 & $104 \pm 16$	& $67 \pm 5$	        & $63 \pm 5$	& $\mathbf{56 \pm 4} $ \\
    5000 & $110 \pm 17$	& $76 \pm 4$	        & $67 \pm 8$	& $\mathbf{64 \pm 4} $ \\
    \hline
    Min $k=25$ & \\
    \hline
    500  & $\mathbf{42} \pm 32$ & $57 \pm  6$	        & $53 \pm 16$           & $52 \pm \mathbf{5} $ \\
    2500 & $\mathbf{44} \pm 32$	& $53 \pm \mathbf{4}$	& $48 \pm 10$	        & $47 \pm 6 $ \\
    5000 & $53 \pm 21$	        & $60 \pm 5$	        & $\mathbf{48} \pm 11$	& $54 \pm \mathbf{3} $ \\
    \hline
    Max $k=25$ & \\
    \hline
    500  & $126 \pm 7$ & $73 \pm 3$	            & $70 \pm 3$    &  $\mathbf{63 \pm 3} $ \\
    2500 & $117 \pm 6$	& $70 \pm 3$	        & $65 \pm 3$	& $\mathbf{60 \pm 3} $ \\
    5000 & $125 \pm 5$	& $80 \pm 2$	        & $72 \pm 2$	& $\mathbf{67 \pm 2} $ \\
\end{tabular}
\caption{\textbf{MC optimization provides consistently superior acquisition scores over built-in optimizers.} Top rows: The mean and standard deviation of all 1250 chosen acquisition samples for the four methods on DNOs trained with 500, 2500, and 5000 samples. Middle rows: The mean and standard deviation of the best acquisition samples for the 25 experiments. Bottom rows: The mean and standard deviation of the worst chosen acquisition samples (i.e. the least optimal sample of the batch). All values are $-\log_{10}(a(\mathbf{x}))$. Best values are bold.}
\label{tab:scores}
\end{table}

\section{DNO \& Likelihood-Weighted Sampling Implementation Tips} \label{app:Tips}
\ethan{While many of these tips are general to neural network implementation, we reiterate them here to assist others in reproducing these results and applying the method to new stochastic problems.}

\ethan{\begin{enumerate}
    \item The input function to the DNO should be scaled to vary within values of -1 and 1.
    \item The quantity of interest, or output, should also be scaled and normalized to values between -1 and 1.
    \item Users should ensure that the DNO is indeed fitting the outputs. This is typically observed with final training errors that are multiple orders of magnitude below the initial training error at epoch 0. To adjust, increase training epochs, layers, and/or width of neurons. Otherwise, the scaling above may have not been appropriately performed.
    \item While we did not experience problems, the weights, $w(\mathbf{x})$, could be negatively affected by the output PDF, $p_{\mu}(\mu)$, in pathological cases. Monitoring the output PDF and its impact on the weights is recommended. We also stress that our approach assumes that extreme events are also rare, as this approach brings attention to the tails. If extremes are not rare, they are not contained within the tails.
    \item We found consecutive training and evaluation runs of \textsc{Tensorflow} with the same kernel resulted in slowing speeds. To combat this, after a round of training, evaluation, and determination of the next acquisition samples are completed, \textsc{Python} is closed and a new instance is loaded for the next step.
\end{enumerate}}

\end{document}